\pdfoutput=1

\documentclass[11pt]{article}

\usepackage{acl}

\usepackage{times}
\usepackage{booktabs}
\usepackage{amsmath}
\usepackage{latexsym}
\usepackage{hyperref}
\usepackage{graphicx}
\usepackage{scalefnt}

\usepackage[T1]{fontenc}

\usepackage[utf8]{inputenc}

\usepackage{microtype}

\newcommand{\titletext}{Few-shot Controllable Style Transfer for Low-Resource Multilingual Settings}

\title{Few-shot Controllable Style Transfer for \\ Low-Resource Multilingual Settings}

\author{\bf Kalpesh Krishna$^{\spadesuit\,*}$ \quad Deepak Nathani$^\diamondsuit$ \quad Xavier Garcia$^\diamondsuit$ \\ \bf \quad Bidisha Samanta$^\diamondsuit$ \quad Partha Talukdar$^\diamondsuit$ \\\\ $^\spadesuit$University of Massachusetts Amherst, $^\diamondsuit$Google Research \\ \texttt{kalpesh@cs.umass.edu}\\ \texttt{\{xgarcia, dnathani, bidishasamanta, partha\}@google.com}}

\newcommand{\namedref}[2]{\hyperref[#2]{#1~\ref*{#2}}}

\newcommand{\relacc}{r-\textsc{acc}}
\newcommand{\absacc}{a-\textsc{acc}}
\newcommand{\acc}{\textsc{acc}}

\newcommand{\ur}{\textsc{ur}}
\newcommand{\urindic}{\textsc{ur-indic}}
\newcommand{\diffur}{\textsc{diffur}}
\newcommand{\diffurindic}{\textsc{diffur-indic}}

\newcommand{\exemp}{\textsc{diffur-mlt}}
\newcommand{\relagg}{r-\textsc{agg}}
\newcommand{\absagg}{a-\textsc{agg}}
\newcommand{\agg}{\textsc{agg}}
\newcommand{\simmetric}{\textsc{sim}}
\newcommand{\lang}{\textsc{lang}}
\newcommand{\copymetric}{\textsc{copy}}
\newcommand{\stylechange}{\textsc{incr}}
\newcommand{\stylecalib}{\textsc{calib}}
\newcommand{\fmetric}{1-g}
\newcommand{\stylecalibinp}{\textsc{c-in}}
\newcommand{\lambdamax}{\lambda_\text{max}}

\newcommand{\sectionref}[1]{\namedref{Section}{#1}}
\newcommand{\tableref}[1]{\namedref{Table}{#1}}
\newcommand{\figureref}[1]{\namedref{Figure}{#1}}
\newcommand{\appendixref}[1]{\namedref{Appendix}{#1}}

\newcommand\blfootnote[1]{%
  \begingroup
  \renewcommand\thefootnote{}\footnote{#1}%
  \addtocounter{footnote}{-1}%
  \endgroup
}

\begin{document}
\maketitle
\begin{abstract}

Style transfer is the task of rewriting a sentence into a target style while approximately preserving content. While most prior literature assumes access to a large style-labelled corpus, recent work~\citep{riley-etal-2021-textsettr} has attempted ``few-shot'' style transfer using just 3-10 sentences at inference for style extraction. In this work, we study a relevant low-resource setting: style transfer for languages where no style-labelled corpora are available. We notice that existing few-shot methods perform this task poorly, often copying inputs \emph{verbatim}.

We push the state-of-the-art for few-shot style transfer with a new method modeling the stylistic difference between paraphrases. When compared to prior work, our model achieves 2-3x better performance in formality transfer and code-mixing addition across seven languages. Moreover, our method is better at controlling the style transfer magnitude using an input scalar knob. We report promising qualitative results for several attribute transfer tasks (sentiment transfer, simplification, gender neutralization, text anonymization) all \emph{without retraining the model}. Finally, we find model evaluation to be difficult due to the lack of datasets and metrics for many languages. To facilitate future research we crowdsource formality annotations for 4000 sentence pairs in four Indic languages, and use this data to design our automatic evaluations.\footnote{Resources accompanying this project will be added to our project page: \url{https://martiansideofthemoon.github.io/2022/03/03/acl22.html}.} 
\blfootnote{*Work done during a Google Research India internship.}

\end{abstract}
\section{Introduction}

\begin{figure}[t]
    \centering
    \includegraphics[width=0.48\textwidth]{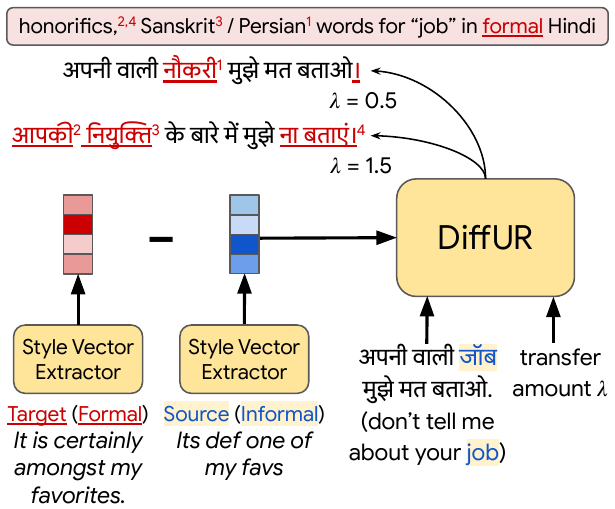}
    \caption{An illustration of our few-shot style transfer system during inference. Our model extracts style vectors from exemplar English sentences as input (in this case formal/informal sentences) and uses their vector difference to guide style transfer in other languages (Hindi). $\lambda$ is used to control the magnitude of transfer: in this example our model produces more high Sanskrit words \& honorifics (more formal) with higher $\lambda$.}
    \vspace{-0.15in}
    \label{fig:diffur-model-inference}
\end{figure}

Style transfer is a natural language generation task in which input sentences need to be re-written into a target style, while preserving semantics. It has many applications such as writing assistance~\citep{heidorn2000intelligent}, controlling generation for attributes like simplicity, formality or persuasion~\citep{xu2015problems,smith2020controlling,niu2020controlling}, data augmentation~\citep{xie2019unsupervised, lee2021neural}, and author obfuscation~\citep{shetty2018a4nt}.

Most prior work either assumes access to supervised data with parallel sentences between the two styles~\citep{jhamtani2017shakespearizing}, or access to a large corpus of unpaired sentences with style labels~\citep{prabhumoye-etal-2018-style,subramanian2018multiple}. Models built are style-specific and cannot generalize to new styles during inference, which is needed for applications like real-time adaptation to a user's style in a dialog or writing application. Moreover, access to a large unpaired corpus with style labels is a strong assumption. Most standard ``unpaired'' style transfer datasets have been carefully curated~\citep{shen2017style} or were originally parallel~\citep{xu-etal-2012-paraphrasing,rao-tetreault-2018-dear}. This is especially relevant in settings outside English, where NLP tools and labelled datasets are largely underdeveloped~\citep{joshi-etal-2020-state}. In this work, we take the first steps studying style transfer in seven languages\footnote{Indic \texttt{(hi,bn,kn,gu,te}), Spanish, Swahili.} with nearly 1.5 billion speakers in total. Since no training data exists for these languages, we analyzed the current state-of-the-art in few-shot multilingual style transfer, the Universal Rewriter (\ur) from~\citet{garcia2021towards}. Unfortunately, we find it often copies the inputs verbatim (\sectionref{sec:universal-rewriter-shortcomings}), \emph{without changing their style}.


We propose a simple inference-time trick of style-controlled translation through English, which improves the \ur~output diversity~(\sectionref{sec:bt-at-inference}). To further boost performance we propose \diffur,\footnote{``\textbf{Diff}erence \textbf{U}niversal \textbf{R}ewriter'', pronounced as \emph{differ}.} a novel algorithm using the recent finding that paraphrasing leads to stylistic changes~\citep{krishna-etal-2020-reformulating}. \diffur~extracts edit vectors from paraphrase pairs, which are used to condition and train the model (\figureref{fig:diffur-model}). On formality transfer and code-mixing addition, our best performing \diffur~variant significantly outperforms \ur~across all languages (by 2-3x) using automatic \& human evaluation. Besides better rewriting, our system is better able to control the style transfer magnitude (\figureref{fig:diffur-model-inference}). A scalar knob ($\lambda$) can be adjusted to make the output text reflect the target style (provided by exemplars)  more or less. We also observe promising qualitative results in several attribute transfer directions (\sectionref{sec:analysis}) including sentiment transfer, simplification, gender neutralization and text anonymization, all \emph{without retraining the model} and using just 3-10 examples at inference.

Finally, we found it hard to precisely evaluate models due to the lack of evaluation datasets and style classifiers (often used as metrics) for many languages. To facilitate further research in Indic formality transfer, we crowdsource formality annotations for 4000 sentence pairs in four Indic languages (\sectionref{sec:meta-eval-dataset}), and use this dataset to design the automatic evaluation suite (\sectionref{sec:evaluation}).

\noindent In summary, our contributions provide an end-to-end recipe for developing and evaluating style transfer models and evaluation in a low-resource setting.

\section{Related Work}

\noindent \textbf{Few-shot methods} are a recent development in English style transfer, with prior work using variational autoencoders~\citep{pmlr-v119-xu20a}, or prompting large pretrained language models at inference~\citep{reif2021recipe}. Most related is the state-of-the-art TextSETTR model from~\citet{riley-etal-2021-textsettr}, who use a neural style encoder to map exemplar sentences to a vector used to guide generation. To train this encoder, they use the idea that adjacent sentences in a document have a similar style. Recently, the \textbf{Universal Rewriter}~\citep{garcia2021towards} extended TextSETTR to 101 languages, developing a joint model for translation, few-shot style transfer and stylized translation. This model is the only prior few-shot system we found outside English, and our main baseline. We discuss its shortcomings in \sectionref{sec:universal-rewriter-shortcomings}, and propose fixes in \sectionref{sec:models}.

\noindent \textbf{Multilingual style transfer} is mostly unexplored in prior work: a 35 paper survey by~\citet{briakou-etal-2021-ola} found only one work in Chinese, Russian, Latvian, Estonian, French. They further introduced XFORMAL, the first formality transfer \emph{evaluation} dataset in French, Brazilian Portugese and Italian.\footnote{We do not use this data since it does not cover Indian languages, and due to Yahoo! L6 corpus restrictions for industry researchers (confirmed via author correspondence).} To the best of our knowledge, we are the first to study style \emph{transfer} for the languages we consider. More related work from Hindi linguistics and on style transfer control is provided in \appendixref{appendix:more-related-work}.



\section{The Universal Rewriter (\ur) model}
\label{sec:universal-rewriter-modeling}

We will start by discussing the Universal Rewriter (\ur) model from~\citet{garcia2021towards}, upon which our proposed \diffur~model is built. At a high level, the \ur~model extracts a style vector $\mathbf{s}$ from an exemplar sentence $e$, which reflects the desired target style. This style vector is used to style transfer an input sentence $x$. Concretely, consider $f_\text{enc}, f_\text{dec}$ to be encoder \& decoder Transformers initialized with mT5~\citep{xue-etal-2021-mt5}, which are composed to form the model $f_\text{ur}$. The \ur~model extracts the style vector using the encoder representation of a special \texttt{[CLS]} token prepended to $e$, and adds it to the input $x$ representations for style transfer,
\begin{align*}
f_\text{style}(e) = \mathbf{s} &= f_\text{enc}(\texttt{[CLS]} \oplus e)[0] \\
    f_\text{ur}(x, \mathbf{s}) &= f_\text{dec}(f_\text{enc}(x) + \mathbf{s})
\end{align*}
where $\oplus$ is string concatenation, $+$ vector addition. $f_\text{ur}$ is trained using the following objectives, 

\vspace{0.05in}

\noindent \textbf{Learning Style Transfer by Exemplar-driven Denoising}: To learn a style extractor, the Universal Rewriter uses the idea that two non-overlapping spans of text in the same document are likely to have the same style. Concretely, let $x_1$ and $x_2$ be two non-overlapping spans. Style extracted from one span ($x_1$) is used to denoise the other ($x_2$),
\begin{align*}
    \bar{x}_2 &= f_\text{ur}(\text{noise}(x_2), f_\text{style}(x_1)) \\
    \mathcal{L}_\text{denoise} &= \mathcal{L}_\text{CE}(\bar{x}_2, x_2)
\end{align*}

where $\mathcal{L}_\text{CE}$ is the standard next-word prediction cross entropy loss function and noise($\cdot$) refers to 20-60\% random token dropping and token replacement. This objective is used on the mC4 dataset~\citep{xue-etal-2021-mt5} with 101 languages. To build a general-purpose rewriter which can do translation as well as style transfer, the model is \textbf{additionally trained on two objectives}: (1) supervised machine translation using the OPUS-100 parallel dataset~\citep{zhang-etal-2020-improving}, and (2) a self-supervised objective to learn effective style-controlled translation; more details in \appendixref{sec:details-ur-model}.

\vspace{0.05in}

\noindent During \textbf{inference} (\figureref{fig:diffur-model-inference}), consider an input sentence $x$ and a transformation from style $A$ to $B$ (say \emph{informal} to \emph{formal}). Let $S_A, S_B$ to be exemplar sentences in each of the styles (typically 3-10 sentences). The output $y$ is computed as,
\begin{align*}
    \mathbf{s}_A &= \frac{1}{|S_A|} \sum_{y \in S_A} f_\text{style}(y)\\
    \mathbf{s}_B &= \frac{1}{|S_B|} \sum_{y \in S_B} f_\text{style}(y)\\
    y &= f_\text{ur}(x, \lambda (\mathbf{s}_B - \mathbf{s}_A))
\end{align*}

where $\lambda$ acts as a control knob to determine the magnitude of style transfer, and the vector subtraction helps remove confounding style information.\footnote{\citet{garcia2021towards} also recommend adding the style vectors from the input sentence $x$, but we found this increased the amount of verbatim copying and led to poor performance.}

\begin{figure*}[t!]
    \centering
    \includegraphics[width=0.95\textwidth]{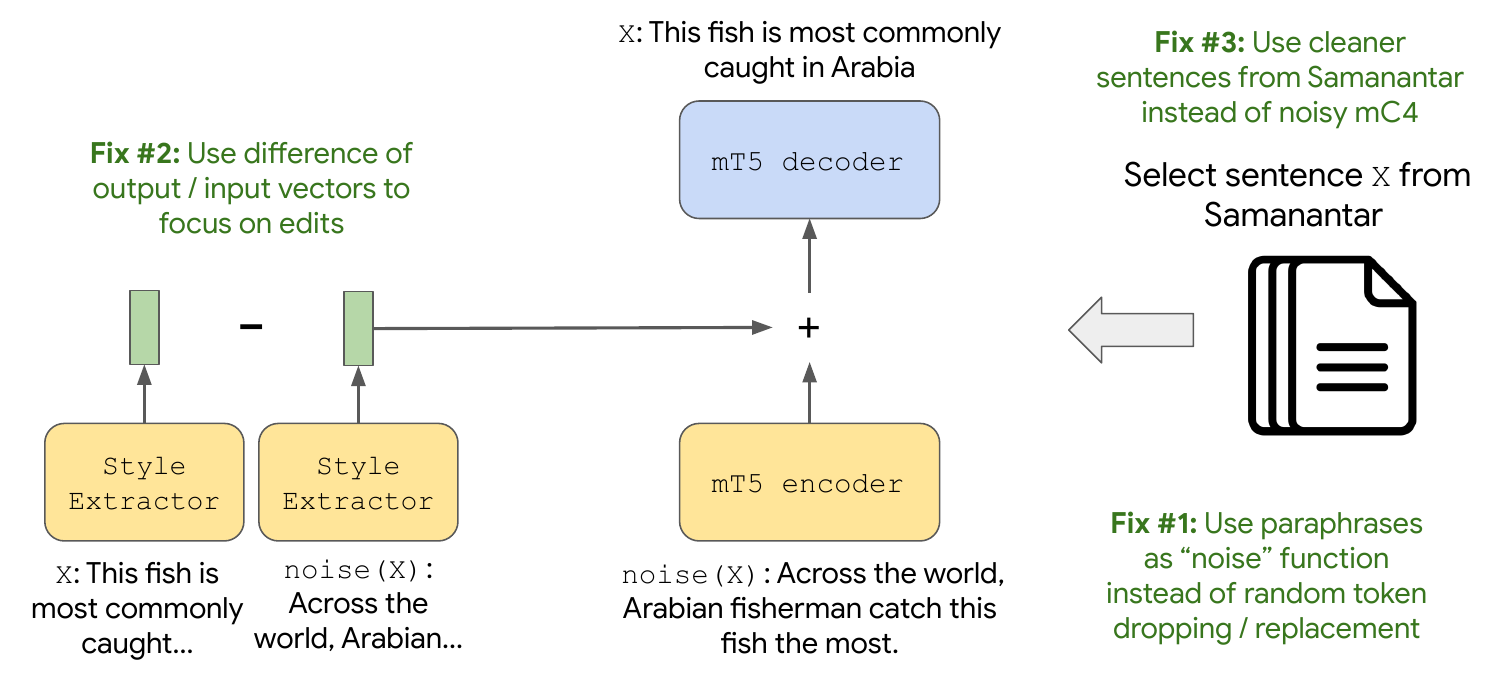}
    \vspace{-0.1in}
    \caption{The \diffur~approach (\sectionref{sec:diffur-model}), with fixes to the shortcomings of the Universal Rewriter approach (\sectionref{sec:universal-rewriter-shortcomings}) shown. Sentences are noised using paraphrasing, the style vector difference between the paraphrase \& original sentence (``edit vector'') is used to control denoising. See \figureref{fig:diffur-model-inference} for the inference-time process.}
    \vspace{-0.15in}
    \label{fig:diffur-model}
\end{figure*}

\subsection{Shortcomings of the Universal Rewriter}
\label{sec:universal-rewriter-shortcomings}

We experimented with the \ur~model on Hindi formality transfer, and noticed poor performance. We noticed that \ur~has a \textbf{strong tendency to copy sentences \emph{verbatim}} --- 45.5\% outputs were copied exactly from the input (and hence not style transferred) for the best performing value of $\lambda$. The copying increase for smaller $\lambda$, making magnitude control harder. We identify the following issues:

\vspace{0.02in}

\noindent 1. \textbf{Random token noise leads to unnatural inputs \& transformations:} The Universal Rewriter uses 20-60\% uniformly random token dropping or replacement to noise inputs, which leads to ungrammatical inputs during training. We hypothesize models tend to learn grammatical error correction, which encourages verbatim copying during inference where fluent inputs are used and no error correction is needed. Moreover, token-level noise does not differentiate between content or function words, and cannot do syntactic changes like content reordering~\citep{goyal2020neural}. Too much noise could distort semantics and encourage hallucination, whereas too little will encourage copying. 

\vspace{0.02in}

\noindent 2. \textbf{Style vectors may not capture the precise style transformation:} The Universal Rewriter extracts the style vector from a single sentence during training, which is a mismatch from the inference where a \emph{difference} between vectors is taken. Without taking vector differences at inference, we observe semantic preservation and overall performance of the \ur~model is much lower.\footnote{This difference possibly helps remove confounding information (like semantic properties, other styles) and focus on the specific style transformation. Since two spans in the same document will share aspects like article topic / subject along with style, we expect these semantic properties will confound the style vector space obtained after the \ur~training.}

\vspace{0.02in}

\noindent 3. \textbf{mC4 is noisy}: On reading training data samples, we noticed noisy samples with severe language identification errors in the Hindi subset of mC4. This has also been observed recently in~\citet{caswell2021quality}, who audit 100 sentences in each language, and report 50\% sentences in Marathi and 20\% sentences in Hindi have the wrong language.

\vspace{0.02in}

\noindent 4. \textbf{No translation data for several languages:} We notice worse performance for languages which did not get parallel translation data (for the translation objective in \sectionref{sec:universal-rewriter-modeling}). In \tableref{tab:formality-eval-test-multi} we see \ur~gets a score\footnote{Using the \relagg~style transfer metric from \sectionref{sec:aggregation-overall-style-transfer}.} of 30.4 for Hindi and Bengali, languages for which it got translation data. However, the scores are lower for Kannada, Telugu \& Gujarati (25.5, 22.8, 23.7), for which no translation data was used. We hypothesize translation data encourages learning language-agnostic semantic representations needed for translation from the given language, which in-turn improves style transfer.
\section{Our Models}
\label{sec:models}

\subsection{Style-Controlled Backtranslation (+ \textsc{bt})}
\label{sec:bt-at-inference}

While the Universal Rewriter model has a strong tendency to exactly copy input sentences while rewriting sentences in the same language (\sectionref{sec:universal-rewriter-shortcomings}), we found it is an effective style-controlled \emph{translation} system. This motivates a simple \textbf{inference-time} trick to improve model outputs and reduce copying --- translate sentences to English (\texttt{en}) in a style-agnostic manner with a zero style vector $\mathbf{0}$, and translate back into the source language (\texttt{lx}) with stylistic control.
\begin{align*}
    x^{\texttt{en}} &= f_\text{ur}(\texttt{en} \oplus x, \mathbf{0}) \\
    \bar{x} &= f_\text{ur}(\texttt{lx} \oplus x^{\texttt{en}}, \lambda (\mathbf{s}_B - \mathbf{s}_A))
\end{align*}

where $x$ is the input sentence, $\mathbf{s}_A, \mathbf{s}_B$ are the styles vectors we want to transfer between, \texttt{en, lx} are language codes prepended to indicate the output language (\appendixref{sec:details-ur-model}). Prior work has shown that backtranslation is effective for paraphrasing~\citep{wieting-gimpel-2018-paranmt,iyyer-etal-2018-adversarial} and style transfer~\citep{prabhumoye-etal-2018-style}.

\subsection{Using Paraphrase Vector Differences for Style Transfer (\diffur)}
\label{sec:diffur-model}

While style-controlled backtranslation is an effective strategy, it needs two translation steps. This is 2x slower than \ur, and semantic errors increase with successive translations. To learn effective style transfer systems needing only a single generation step we develop \diffur, a new few-shot style transfer training objective (overview in \figureref{fig:diffur-model}). \diffur~tackles the issues discussed in \sectionref{sec:universal-rewriter-shortcomings} using paraphrases and style vector differences.

\vspace{0.05in}

\noindent \textbf{Paraphrases as a ``noise'' function}: Instead of using random token-level noise (Issue \#1 in \sectionref{sec:universal-rewriter-shortcomings}), we paraphrase sentences to ``noise'' them during training. Paraphrasing modifies the lexical \& syntactic properties of sentences, while preserving fluency and input semantics. Prior work~\citep{krishna-etal-2020-reformulating} has shown that
paraphrasing leads to stylistic changes, and denoising can be considered a style re-insertion process.

To create paraphrases, we backtranslate sentences from the \ur~model\footnote{Specifically, an Indic variant of the \ur~model is used, described in \sectionref{sec:indic-variants}. Note it is not necessary to use \ur~for backtranslation, any good translation model can be used.} with no style control (zero vectors used as style vectors). To increase diversity, we use random sampling in both translation steps, pooling generations obtained using temperature values $[0.4, 0.6, 0.8, 1.0]$. Finally, we discard paraphrase pairs from the training data where the semantic similarity score\footnote{Calculated using LaBSE, discussed in \sectionref{sec:semantic-similarity}.} is outside the range $[0.7, 0.98]$. This removes backtransation errors (score < 0.7), and exact copies (score > 0.98). In \appendixref{appendix:paraphrase-diversity} we confirm that our backtranslated paraphrases are lexically diverse from the input.

\vspace{0.05in}

\noindent \textbf{Using style vector differences for control}: To fix the training / inference mismatch for style extraction (Issue \#2 in \sectionref{sec:universal-rewriter-shortcomings}), we propose using style vector differences between the output and input as the stylistic control. Concretely, let $x$ be an input sentence and $x_\text{para}$ its paraphrase.
\begin{align*}
    \mathbf{s}_\text{diff} &= f_\text{style}(x) - f_\text{style}(x_\text{para})\\
    \bar{x} &= f_\text{ur}(x_\text{para}, \text{stop-grad}(\mathbf{s}_\text{diff})) \\
    \mathcal{L} &= \mathcal{L}_\text{CE}(\bar{x}, x)
\end{align*}

where stop-grad($\cdot$) stops gradient flow through $\mathbf{s}_\text{diff}$, preventing the model from learning to copy $x$ exactly. To ensure $f_\text{style}$ extracts meaningful style representations, we fine-tune a trained $\ur$ model. Vector differences have many advantages,
\begin{enumerate}
\setlength{\itemsep}{0.0pt}
    \item Subtracting style vectors between a sentence and its paraphrase removes confounding features (like semantics) present in the vectors.
    
    \item The vector difference focuses on the precise transformation that is needed to reconstruct the input from its paraphrase.
    
    \item The length of $\mathbf{s}_\text{diff}$ acts as a proxy for the amount of style transfer, which is controlled using $\lambda$ during inference (\sectionref{sec:universal-rewriter-modeling}).
\end{enumerate}
\diffur~is related to neural editor models~\citep{guu2018generating,he2020learning}, where language models are decomposed into a probabilistic space of edit vectors over prototype sentences. We justify the \diffur~design with ablations in \appendixref{appendix:ablation-explanations}.

\subsection{Indic Models (\urindic, \diffurindic)}
\label{sec:indic-variants}

To address the issue of no translation data (Issue \#4 in \sectionref{sec:universal-rewriter-shortcomings}), we train Indic variants of our models. We replace the OPUS translation data used for training the Universal Rewriter (\sectionref{sec:universal-rewriter-modeling}) with Samanantar~\citep{ramesh2021samanantar}, which is the largest publicly available parallel translation corpus for 11 Indic languages. We call these variants \urindic~and \diffurindic. This process significantly up-samples the parallel data seen between English / Indic languages, and gives us better performance (\tableref{tab:formality-eval-test-multi}) and lower copy rates, especially for languages with no OPUS translation data.

\subsection{Multitask Learning (\exemp)}
\label{sec:multitask}

One issue with our \diffurindic~setup is usage of a stop-grad($\cdot$) to avoid verbatim copying from the input. This prevents gradient flow into the style extractor $f_\text{style}$, and as we see in \appendixref{appendix:analysis}, a degradation of the style vector space. To prevent this we simply multi-task between the exemplar-driven denoising \ur~objective (\sectionref{sec:universal-rewriter-modeling}) and the \diffur~objective. We initialize the model with the \urindic~checkpoint, and fine-tune it on these two losses together, giving each loss equal weight.

\section{Evaluation}
\label{sec:evaluation}

Automatic evaluation of style transfer is challenging~\citep{pang-2019-towards,mir-etal-2019-evaluating,tikhonov2019style}, and the lack of resources (such as evaluation datasets, style classifiers) make evaluation trickier for Indic languages. To tackle this issue, we first collect a small dataset of formality and semantic similarity annotations in four Indic languages (\sectionref{sec:meta-eval-dataset}). We use this dataset to guide the design of an evaluation suite (\sectionref{sec:accuracy-eval}-\ref{sec:evaluating-control}). Since automatic metrics in generation are imperfect~\citep{celikyilmaz2020evaluation}, we complement our results with human evaluation (\sectionref{sec:human-evaluation}).


\subsection{Indic Formality Transfer Dataset}
\label{sec:meta-eval-dataset}

Since no public datasets exist for formality transfer in Indic languages, it is hard to measure the extent to which automatic metrics (such as style classifiers) are effective. To tackle this issue, we build a dataset of \textbf{1000} sentence pairs in \textbf{each of four Indic languages} (Hindi, Bengali, Kannada, Telugu) with formality and semantic similarity annotations. We first style transfer held-out Samanantar sentences using our \urindic~+~\textsc{bt} model (\sectionref{sec:bt-at-inference}, \ref{sec:indic-variants}) to create sentence pairs with different formality. We then asked three crowdworkers to 1) label the more formal sentence in each pair; 2) rate semantic similarity on a 3-point scale.

Our crowdsourcing is conducted on Task Mate,\footnote{\url{https://taskmate.google.com}} where we hired native speakers from India with at least a high school education and 90\% approval rating on the platform. To ensure crowdworkers understood ``formality'', we provided instructions following advice from professional Indian linguists, and asked two qualification questions in their native language. More details (agreement, compensation, instructions) are provided in \appendixref{appendix:crowdsourcing-setup-details}.

\subsection{Transfer Accuracy (\relacc, \absacc)}
\label{sec:accuracy-eval}

Our first metric checks whether the output sentence reflects the target style. This is measured by an external classifier's predictions on system outputs. We use two variants of transfer accuracy: (1) Relative Accuracy (\relacc): does the target style classifier score the output sentence \emph{higher} than the input sentence? (2) Absolute Accuracy (\absacc): does the classifier score the output \emph{higher} than 0.5?

\noindent \textbf{Building multilingual classifiers}: Unfortunately, no large style classification datasets exist for most languages, preventing us from building classifiers from scratch. We resort to zero-shot cross lingual transfer techniques~\citep{conneau2019cross}, where large multilingual pretrained models are first fine-tuned on English classification data, and then applied to other languages at inference. We experiment with three such techniques, and find MAD-X classifiers with language adapters~\citep{pfeiffer-etal-2020-mad} have the highest accuracy of 81\% on our Hindi data from \sectionref{sec:meta-eval-dataset}. However, MAD-X classifiers were only available for Hindi, so we use the next best XLM RoBERTa-base~\citep{conneau-etal-2020-unsupervised} for other languages, which has 75\%-82\% accuracy on annotated data; details in \appendixref{appendix:classifier}.

\subsection{Semantic Similarity (\simmetric)}
\label{sec:semantic-similarity}

Our second evaluation criteria is semantic similarity between the input and output. Following recent recommendations~\citep{marie-etal-2021-scientific, krishna-etal-2020-reformulating}, we avoid $n$-gram overlap metrics like BLEU~\citep{papineni-etal-2002-bleu}. Instead, we use LaBSE~\citep{feng2020language}, a language-agnostic semantic similarity model based on multilingual BERT~\citep{devlin-etal-2019-bert}. LaBSE supports 109 languages, and is the only similarity model we found supporting all the Indic languages in this work. We also observed LaBSE had greater correlation with our annotated data (\sectionref{sec:meta-eval-dataset}) compared to alternatives; details in \appendixref{appendix:semantic-similarity}.

Qualitatively, we found that sentence pairs with LaBSE scores lower than 0.6 were almost never paraphrases. To avoid rewarding partial credit for low LaBSE scores, we use a hard threshold\footnote{ Roughly 73\% pairs annotated as paraphrases (from dataset in \sectionref{sec:meta-eval-dataset}) had $L > 0.75$. We experiment with different values of $L$ in  \appendixref{sec:eval-different-labse} and notice similar trends.} ($L=0.75$) to determine whether pairs are paraphrases,
\begin{align*}
    \text{\simmetric}(x, y') = 1 \text { if } \left\{ \text{LaBSE}(x, y') > L \right\} \text{ else } 0
\end{align*}

\subsection{Other Metrics (\textsc{lang}, \copymetric, \fmetric)}
\label{sec:langid}

Additionally, we measure whether the input and output sentences are in the same language (\textsc{lang}), the fraction of outputs copied verbatim from the input (\copymetric), and the 1-gram overlap between input / output (\fmetric). High \textsc{lang} and low \copymetric~/ \fmetric~(more diversity) is better; details in \appendixref{appendix:other-individual-details}.

\begin{table*}[t!]
\small
\begin{center}
\begin{tabular}{ lrrrrrrrrrrrr } 
 \toprule
 Model & \multicolumn{2}{c}{Hindi} & \multicolumn{2}{c}{Bengali} & \multicolumn{2}{c}{Kannada} & \multicolumn{2}{c}{Telugu} & \multicolumn{2}{c}{Gujarati} &  \\
 & \relagg & \absagg  & \relagg & \absagg  & \relagg & \absagg  & \relagg & \absagg  & \relagg & \absagg\\
 \midrule
 \ur~\shortcite{garcia2021towards} & 30.4 & 10.4 & 30.4 & 7.2 & 25.5 & 8.0 & 22.8 & 8.4 & 23.7 & 5.0\\
  \urindic & 58.3 & 18.6 & 65.5 & 22.3 & 61.3 & 17.8 & 59.8 & 19.9 & 54.0 & 10.7 \\
  \midrule
 \ur~+ \textsc{bt} & 54.2 & 17.8 & 55.6 & 16.9 & 39.8 & 11.9 & 38.4 & 11.6 & 46.3 & 10.4 \\
 \urindic~+ \textsc{bt} & 60.0 & 22.2 & 61.1 & 22.0 & 59.2 & 21.0 & 56.8 & 22.2 & 57.7 & 16.8 \\
 \midrule
  \diffur & 71.1 & 22.9 & 72.7 & 25.2 & 69.2 & 29.1 & 69.4 & 27.1 & 0.4 & 0.2 \\
 \diffurindic & 72.6 & 24.0 & 75.4 & 24.3 & 73.1 & 29.3 & 71.0 & 27.1 & 36.0 & 13.0 \\
 \exemp & \textbf{78.1} & \textbf{32.2} & \textbf{80.0} & \textbf{35.0} & \textbf{80.4} & \textbf{39.4} & \textbf{79.8} & \textbf{37.9} & \textbf{75.0} & \textbf{33.1} \\
\bottomrule
\end{tabular}
\end{center}
\vspace{-0.1in}
\caption{Automatic evaluation of formality transfer in Indic languages. Note each proposed method (\textsc{*-indic}, +\textsc{bt}, \diffur) improves performance (\textsc{agg} defined in \sectionref{sec:aggregation-overall-style-transfer}), with a combination (\exemp) doing best.}
\vspace{-0.1in}
\label{tab:formality-eval-test-multi}
\end{table*}

\subsection{Aggregated Score (\relagg, \absagg)}
\label{sec:aggregation-overall-style-transfer}

To get a sense of overall system performance, we combine individual metrics into one score. Similar to~\citet{krishna-etal-2020-reformulating} we aggregate metrics as,
\begin{align*}
    \text{\agg}(x, y') &= \text{\acc}(x, y') \cdot \text{\simmetric}(x, y') \cdot \text{\textsc{lang}}(y') \\
    \text{\agg}(\mathcal{D}) &= \frac{1}{|\mathcal{D}|}\sum_{x, y' \in \mathcal{D}} \text{\agg}(x, y')
\end{align*}

\noindent Where $(x, y')$ are input-output pairs, and $\mathcal{D}$ is the test corpus. Since each of our individual metrics can only take values 0 or 1 at an instance level, our aggregation acts like a Boolean AND operation. In other words, we are measuring the fraction of outputs which \emph{simultaneously} transfer style, have a semantic similarity of at least $L$~(our threshold in \sectionref{sec:semantic-similarity}), and have the same language as the input. Depending on the variant of \acc~(relative / absolute), we can derive \relagg~/ \absagg.

\subsection{Evaluating Control (\stylecalib)}
\label{sec:evaluating-control}

An ideal system should not only be able to style transfer sentences, but also control the \emph{magnitude} of style transfer using the scalar input $\lambda$. To evaluate this, for every system we first determine a $\lambdamax$ value and let $[0, \lambdamax]$ be the range of control values. While in our setup $\lambda$ is an unbounded scalar, we noticed high values of $\lambda$ significantly perturb semantics (also noted in \citealp{garcia2021towards}), with systems outputting style-specific $n$-grams unfaithful to the output. We choose $\lambdamax$ to be the largest $\lambda$ from the list $[0.5, 1.0, 1.5, 2.0, 2.5, 3.0]$ whose outputs have an average semantic similarity score (\textsc{sim}, \sectionref{sec:semantic-similarity}) of at least 0.75\footnote{This threshold is identical to the value chosen for paraphrase similarity in \sectionref{sec:semantic-similarity}. We experiment with more/less conservative thresholds in \appendixref{sec:eval-different-labse}.} with the validation set inputs. For each system we take three evenly spaced $\lambda$ values in its control range, denoted as $\Lambda = [\frac{1}{3}\lambdamax$, $\frac{2}{3}\lambdamax$, $\lambdamax]$. We then compute the \textbf{style calibration to $\lambda$} (\stylecalib), or how often does increasing $\lambda$ lead to a style score increase? We measure this with a statistic similar to Kendall's $\tau$~\citep{kendall1938new}, counting concordant pairs in $\Lambda$,
\begin{align*}
    \stylecalib(x) &= \frac{1}{n}\sum_{\lambda_b > \lambda_a} \left\{ \text{style}(y_{\lambda_b}) > \text{style}(y_{\lambda_a}) \right\}
\end{align*}

where $x$ is input, \stylecalib($x$) is the average over all possible $n$ ($=3$) pairs of $\lambda$ values $(\lambda_a, \lambda_b)$ in $\Lambda$.

\subsection{Human Evaluation}
\label{sec:human-evaluation}

Automatic metrics are usually insufficient for style transfer evaluation --- according to~\citet{briakou2021review}, 69 / 97 surveyed style transfer papers used human evaluation. We adopt the crowd-sourcing setup from \sectionref{sec:meta-eval-dataset}, which was used to build our formality evaluation datasets. We presented 200 generations from each model and the corresponding inputs in a random order, and asked three crowdworkers two questions about each pair of sentences: (1) which sentence is more formal/code-mixed? (2) how similar are the two sentences in meaning? This lets us evaluate \relacc, \simmetric, \relagg, \stylecalib~with respect to human annotations instead of classifier predictions. More experiment details (inter-annotator agreement, compensation, instructions) are provided in \appendixref{appendix:crowdsourcing-setup-details}.

\begin{table*}[t!]
\small
\begin{center}
\begin{tabular}{ lrrrrrrr|rr } 
 \toprule
 Model & $\lambda$ & \copymetric $(\downarrow)$ & \fmetric $(\downarrow)$ & \lang & \simmetric & \relacc & \absacc & \relagg & \absagg \\
 \midrule
 \ur~\citep{garcia2021towards} & 1.5 & 45.4 & 77.5 & 98.0 & 84.8 & 45.8 & 22.9 & 30.4 & 10.4 \\
 \urindic  & 1.0 & 10.4 & 70.7 & 95.0 & 93.8 & 67.2 & 23.3 & 58.3 & 18.6 \\
 \midrule
 \ur~+ \textsc{bt} & 0.5 & 0.8 & 44.2 & 92.9 & 85.2 & 72.3 & 27.8 & 54.2 & 17.8 \\
 \urindic~+ \textsc{bt} & 1.0 & 1.1 & 49.5 & 95.9 & 85.1 & 76.3 & 33.1 & 60.0 & 22.2 \\
 \midrule
 \diffur & 1.0 & 4.7 & 61.6 & 97.7 & 89.7 & 82.4 & 31.0 & 71.1 & 22.9  \\
 \diffurindic &  1.5 & 5.3 & 63.7 & 98.0 & 91.9 & 81.6 & 30.5 & 72.5 & 23.7 \\
 \exemp & 2.5 & 4.4 & 61.9 & 97.2 & 89.7 & 89.7 & 34.0 & \textbf{78.1} & \textbf{27.5} \\
\bottomrule
\end{tabular}
\end{center}
\vspace{-0.05in}
\caption{Performance by individual metrics for Hindi formality transfer. \exemp~gives best overall performance (\relagg~/ \absagg), with a good trade-off between style accuracy (\acc), semantic similarity (\simmetric), langID score (\lang), and low input copy rates (\copymetric); metrics defined in \sectionref{sec:evaluation}, other language results in \appendixref{appendix:full-results-breakdown}.}
\vspace{-0.05in}
\label{tab:hindi-test-formality-eval-main}
\end{table*}

\section{Main Experiments}
\label{sec:main-experiments}



\subsection{Experimental Setup}
\label{sec:experimental-setup}

\noindent In our experiments, we compare the following models (training details are provided \appendixref{sec:model-training}):
\begin{itemize}
\setlength{\itemsep}{-0.01pt}
    \item \ur: the Universal Rewriter~\citep{garcia2021towards}, which is our main baseline (\sectionref{sec:universal-rewriter-modeling}); 
    \item \diffur: our model with paraphrase vector differences (\sectionref{sec:diffur-model});
    \item \urindic, \diffurindic: Indic~variants of \ur~and \diffur~models~(\sectionref{sec:indic-variants});
    \item \exemp: Multitask training between \urindic~and \diffurindic~(\sectionref{sec:multitask});
    \item +~\textsc{bt}: models with style-controlled backtranslation at inference time (\sectionref{sec:bt-at-inference}).
\end{itemize}

\noindent Our models are \textbf{evaluated} on (1) formality transfer~\citep{rao-tetreault-2018-dear}; (2) code-mixing addition, a task where systems attempt to use English words in non-English sentences, while preserving the original script.\footnote{ \href{https://en.wikipedia.org/wiki/Hinglish}{Hinglish} is common in India, examples in \figureref{fig:code-mixing-exemplars}.} Since we do not have access to any formality evaluation dataset,\footnote{We do not use GYAFC~\citep{rao-tetreault-2018-dear} and XFORMAL~\citep{briakou-etal-2021-ola} due to reasons in footnote 4. Our dataset from \sectionref{sec:meta-eval-dataset} has already been used for classifier selection, and has machine generated sentences.} we hold out 22K sentences from Samanantar in each Indic language for validation / testing. For Swahili / Spanish, we use mC4 / WMT2018 sentences. These sets have similar number of formal / informal sentences, as marked by our formality classifiers (\sectionref{sec:accuracy-eval}), and are transferred to the opposite formality. We re-use the hi/bn formality transfer splits for code-mixing addition, evaluating unidirectional transfer.
 
 \vspace{0.02in}
 
\noindent \textbf{Seven languages} with varying scripts and morphological richness are used for evaluation (\texttt{hi,es,sw,bn,kn,te,gu}). The \ur~model only saw translation data for \texttt{hi,es,bn}, whereas \urindic~sees translation data for all Indic languages (\sectionref{sec:indic-variants}). To test the generalization capability of the \diffur, no Gujarati paraphrase training data for is used. Note that \emph{no paired/unpaired data with style labels is used during training}: models determine the target style at inference using 3-10 exemplars sentences. For few-shot formality transfer, we use the English exemplars from~\citet{garcia2021towards}. We follow their setup and use English exemplars to guide non-English transfer zero-shot. For code-mixing addition, we use Hindi/English code-mixed exemplars in Devanagari (shown in \appendixref{appendix:exemplar-choice}).

\begin{table}[t!]
\footnotesize
\begin{center}
\begin{tabular}{ lrr } 
 \toprule
 Model & \multicolumn{1}{c}{Swahili} & \multicolumn{1}{c}{Spanish} \\
 & \relagg~/ \absagg  & \relagg~/ \absagg  \\
 \midrule
 \ur~\shortcite{garcia2021towards} & 19.9 / 4.8 & 13.4 / ~1.3 \\
 \ur, \textsc{bt} & 13.7 / 3.4 & 33.3 / ~5.8  \\
 \exemp & \textbf{32.2} / \textbf{7.2} & \textbf{46.5} / \textbf{16.5} \\
\bottomrule
\end{tabular}
\end{center}
\vspace{-0.1in}
\caption{Automatic evaluation of formality transfer in Swahili and Spanish. \exemp~ performs best.}
\vspace{-0.1in}
\label{tab:formality-evaluation-swes}
\end{table}

\begin{table}[t!]
\small
\begin{center}
\begin{tabular}{ lrrrrr } 
 \toprule
 Model & \acc & \simmetric & \agg & \stylecalib & \textsc{c-in} \\
 \midrule
 \ur~\shortcite{garcia2021towards} & 29.5 & \textbf{87.2} & 23.2 & - & - \\
 \urindic & 46.5 & 85.3 & 40.8 & 35.7 & 43.0 \\
\midrule
 \ur~+ \textsc{bt} & 57.5 & 71.2 & 42.9 & - & -\\
 \urindic~+ \textsc{bt} & 65.0 & 77.8 & 52.4 & 24.0 & 40.3 \\
 \midrule 
 \diffur & 64.5 & 80.8 & 52.0 & - & - \\
 \diffurindic & 62.0 & 83.1 & 50.4 & 48.0 & \textbf{54.5} \\
 \exemp & \textbf{70.0} & 80.8 & \textbf{55.6} & \textbf{53.0} & \textbf{54.5} \\
\bottomrule
\end{tabular}
\end{center}
\vspace{-0.1in}
\caption{Human evaluation on Hindi formality transfer, measuring style accuracy (\acc), input similarity (\simmetric), overall score (\agg) and control with $\lambda$ (\stylecalib, \stylecalibinp). Like \tableref{tab:formality-eval-test-multi}, \exemp~performs best.}
\vspace{-0.1in}
\label{tab:hindi-formality-eval-test-human}
\end{table}

\begin{figure*}[t!]
    \centering
    \includegraphics[width=0.37\textwidth]{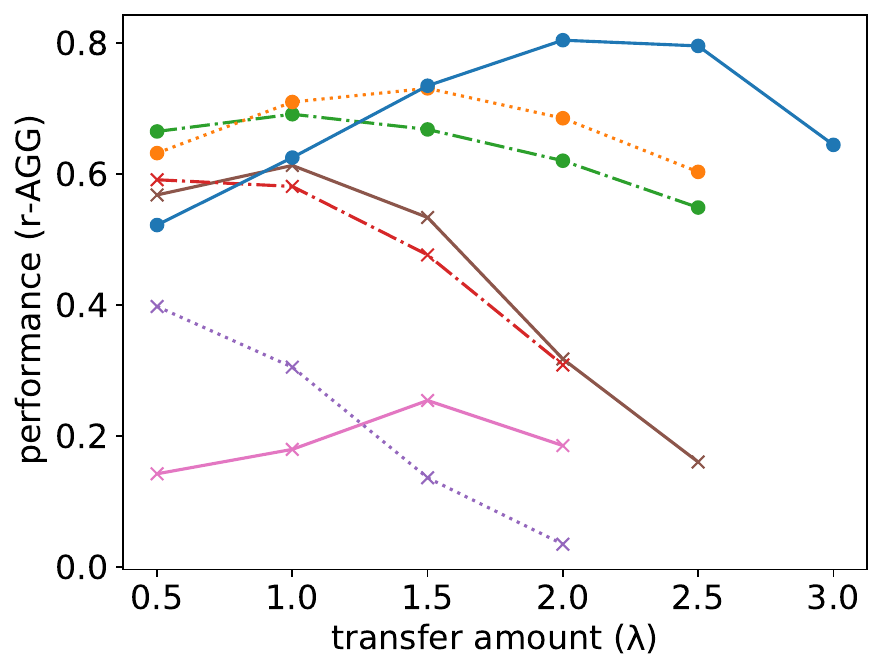}
    \includegraphics[width=0.565\textwidth]{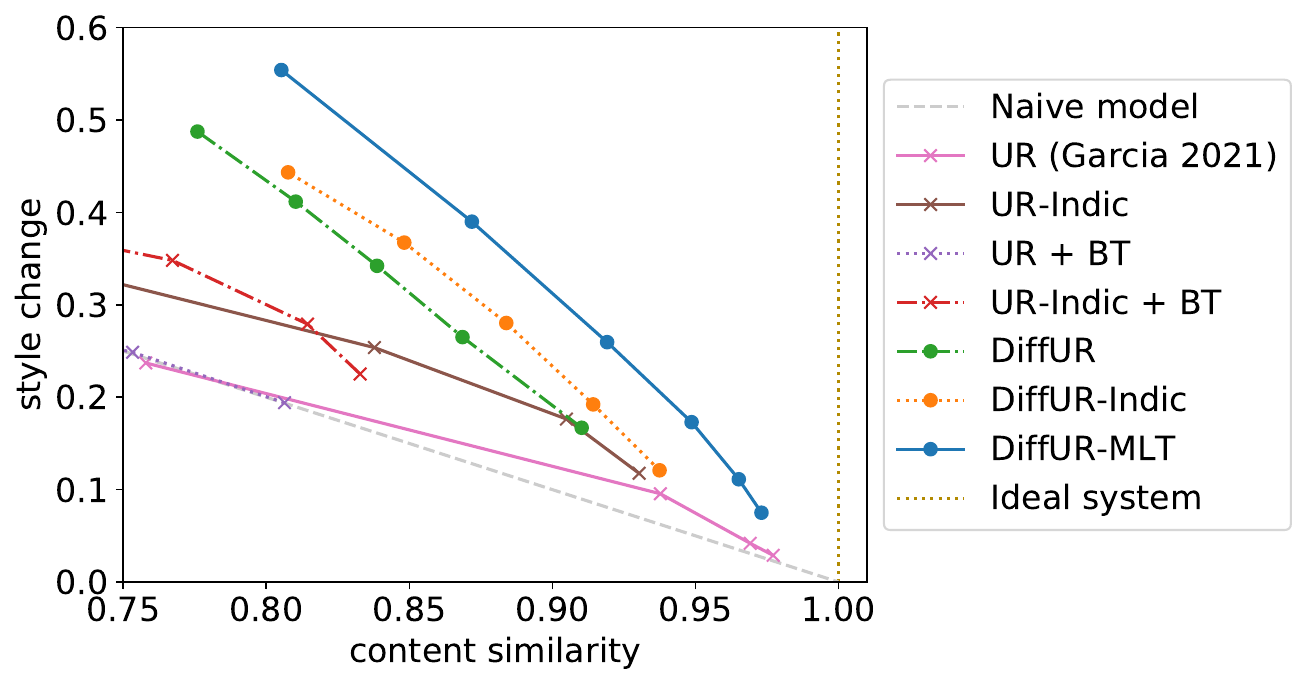}
    \vspace{-0.1in}
    \caption{Variation in Kannada formality transfer with $\lambda$. In the \emph{left} plot, we see \diffur-* models have consistently good overall performance with change in $\lambda$. In the \emph{right} plot, we see the tradeoff between average style change and content similarity as $\lambda$ is varied. Plots (such as \diffur-*) which stretch the Y-axis range, closer to the ideal system ($x=1$) and away from the naive system ($x + y = 1$, akin to naive model in~\citealp{krishna-etal-2020-reformulating}) are better.}
    \label{fig:kannada-variation-across-scales-main}
    \vspace{-0.1in}
\end{figure*}


\subsection{Main Results}
\label{sec:main-results}
\label{sec:analysis}

\noindent \textbf{Each proposed method improves over prior work, \exemp~works best}. We present our automatic evaluation results for formality transfer across languages in \tableref{tab:formality-eval-test-multi}, \tableref{tab:formality-evaluation-swes}. Overall we find that each of our proposed methods (\diffur, \textsc{*-indic}, +\textsc{bt}) helps improve performance over the baseline \textsc{ur} model (71.1, 58.3, 54.2 vs 30.4 \relagg~on Hindi). Combining these ideas with multitask learning (\exemp) gives us the best performance across all languages (78.1 on Hindi). On Gujarati, the \diffurindic~fails to get good performance (36.0 \relagg) since it did not see Gujarati paraphrase data, but this performance is recovered using \exemp~(75.0). In \tableref{tab:hindi-formality-eval-test-human} we see human evaluations support our automatic evaluation for formality transfer. In \tableref{tab:code-mixing-eval-test-human-multi} we perform human evaluation on a subset of models for code-mixing addition and see similar trends, with \exemp~significantly outperforming \ur, \urindic~(41.5 \textsc{agg} vs 3.6, 15.3 on Hindi).

\begin{table}[t!]
\footnotesize
\begin{center}
\begin{tabular}{ lrr } 
 \toprule
 Model & \multicolumn{1}{c}{Hindi} & \multicolumn{1}{c}{Bengali} \\
 & \acc~/ \simmetric~/ \agg  & \acc~/ \simmetric~/ \agg  \\
 \midrule
 \ur~\shortcite{garcia2021towards} & 4.5 / \textbf{93.8} /~~ 3.6 & 0.0 / \textbf{96.4} /~~ 0.0 \\
 \urindic,\textsc{bt} & 18.5 / 79.2 / 15.3 & 18.0 / 68.3 / 12.7 \\
 \exemp,\textsc{bt} & \textbf{62.5} / 69.9 / \textbf{41.5} & \textbf{79.0} / 57.1 / \textbf{43.5} \\
\bottomrule
\end{tabular}
\end{center}
\vspace{-0.1in}
\caption{ Human evaluation on code-mixing addition. \exemp+\textsc{bt} performs best (\agg), giving high style accuracy (\acc). Due to verbatim copying, \ur~\simmetric~score is nearly 100, but \acc~score close to 0.}
\vspace{-0.1in}
\label{tab:code-mixing-eval-test-human-multi}
\end{table}

\begin{table}[t!]
\small
\begin{center}
\begin{tabular}{ lr|lr } 
 \toprule
  Model & \stylecalib & Model & \stylecalib \\
 \midrule
 \ur~\shortcite{garcia2021towards} & 29.2 & \diffur & 64.9 \\
 \urindic & 60.7 & \diffurindic & \textbf{69.6}\\
  \ur~+ \textsc{bt} & 43.4 & \exemp & 69.0   \\
 \urindic~+ \textsc{bt} & 38.7 & \\
\bottomrule
\end{tabular}
\end{center}
\vspace{-0.1in}
\caption{Evaluation of Hindi formality transfer magnitude control using $\lambda$. We find that \diffur-* are best at calibrating style change (\stylecalib) to input $\lambda$ (metrics details in \sectionref{sec:evaluating-control}, more results in \appendixref{appendix:more-control-evals}).}
\vspace{-0.1in}
\label{tab:hindi-formality-eval-test-control}
\end{table}

\begin{figure*}[t]
    \centering
    \includegraphics[width=\textwidth]{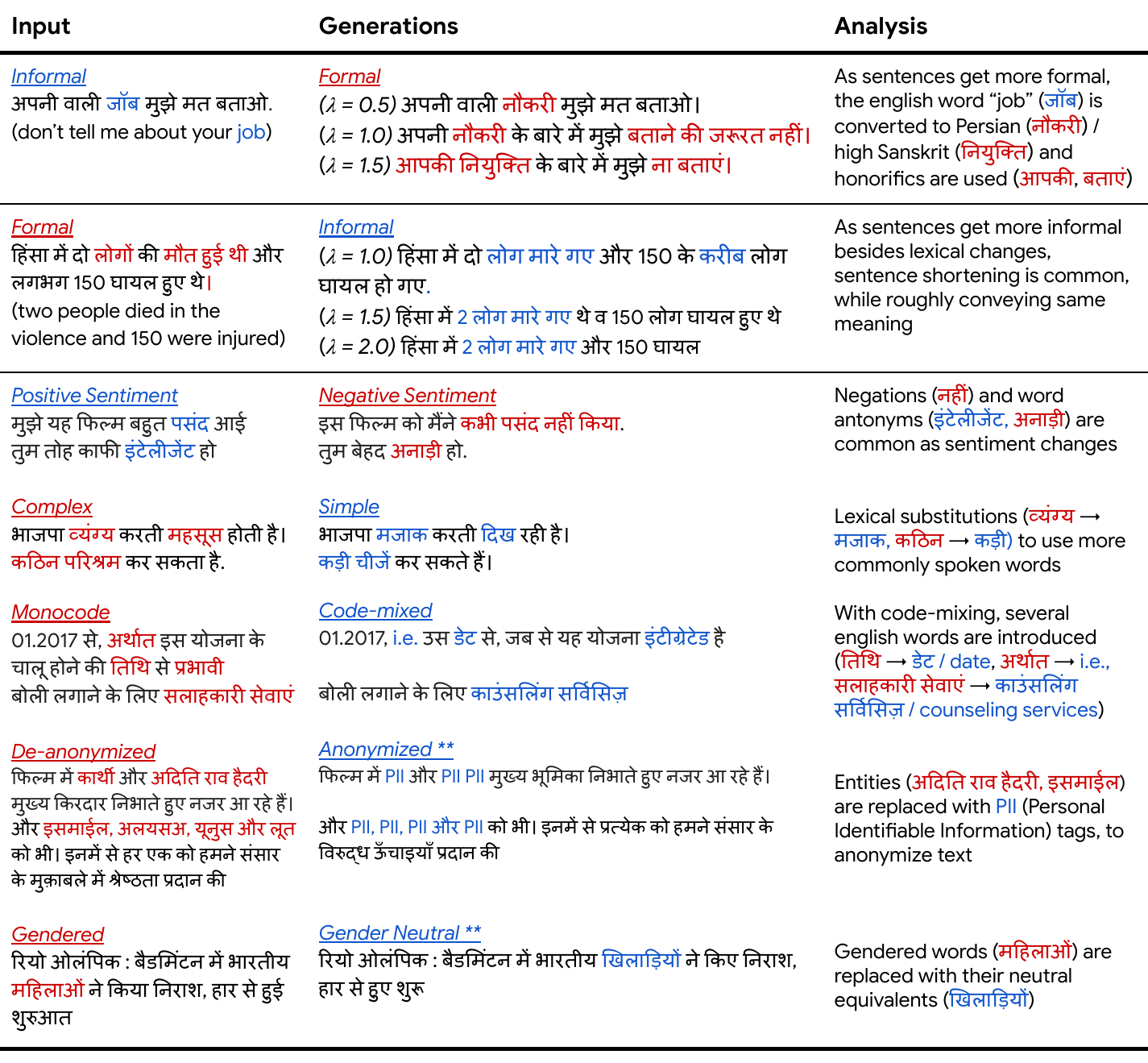}
    \vspace{-0.1in}
    \caption{Outputs and qualitative analysis of our best performing model for several attribute transfer tasks ($\lambda$ is transfer magnitude). We notice lower quality qualitatively for ** marked styles; see \appendixref{appendix:model-outputs} for more outputs.}
    \vspace{-0.1in}
    \label{fig:qualitative}
\end{figure*}




\noindent \textbf{\exemp~and \diffurindic~are best at controlling magnitude of style transfer}: In \tableref{tab:hindi-formality-eval-test-control}, we compare the extent to which models can control the amount of style transfer using $\lambda$. We find that all our proposed methods outperform the \ur~model, which gets only 29.2 \stylecalib. $+\textsc{bt}$ models are not as effective at control (43.4 \stylecalib), while \diffurindic~and \exemp~perform best (69.6, 69.0 \stylecalib). This is graphically illustrated in \figureref{fig:kannada-variation-across-scales-main}. \exemp~performs consistently well across different $\lambda$ values (left plot), and gives a high style change without much drop in content similarity to the input as $\lambda$ is varied (right plot); more control experiments in \appendixref{appendix:more-control-evals}.

 In \tableref{tab:hindi-test-formality-eval-main} we provide a \textbf{breakdown by individual metrics}. In the baseline Hindi \ur~model, we notice high \copymetric~rates (45.4\%), resulting in lower \textsc{acc}~scores. \copymetric~reduces in our proposed models (4.4\% for \exemp), which boosts overall performance. We find the lowest \copymetric~(and lowest \fmetric) for models with +\textsc{bt} (1\%), which is due to two translation steps. However, this lowers semantic similarity (also seen in \tableref{tab:hindi-formality-eval-test-human}) lowering the overall score (60.0 vs 78.1) compared to \exemp. 
 
 In \appendixref{appendix:ablation-studies} we show \textbf{ablations studies} justifying the \diffur~design, decoding scheme, etc. In \appendixref{appendix:full-results-breakdown} we show a breakdown by individual metrics for other languages and plot variations with $\lambda$. We also analyze the style encoder $f_\text{style}$ in \appendixref{appendix:analysis}, finding it is an effective style classifier.


We \textbf{analyze several qualitative outputs} from \exemp~in \figureref{fig:qualitative}. Besides formality transfer and code-mixing addition, we transfer several other attributes: sentiment~\citep{li-etal-2018-delete}, simplicity~\citep{xu2015problems}, anonymity~\citep{anandan2012t} and gender neutrality~\citep{reddy2016obfuscating}. More outputs are provided in \appendixref{appendix:model-outputs}.

\section{Conclusion}

We present a recipe for building \& evaluating controllable few-shot style transfer systems needing only 3-10 style examples at inference, useful in low-resource settings. Our methods outperform prior work in formality transfer \& code-mixing for 7 languages, with promising qualitative results for several other attribute transfer tasks. Future work includes further improving systems for some attributes, and studying style transfer for languages where little / no translation data is available.




\section*{Acknowledgements}

We are very grateful to the Task Mate team (especially Auric Bonifacio Quintana) for their support and helping us crowdsource data and evaluate models on their platform. We thank John Wieting, Timothy Dozat, Manish Gupta, Rajesh Bhatt, Esha Banerjee, Yixiao Song, Marzena Karpinska, Aravindan Raghuveer, Noah Constant, Parker Riley, Andrea Schioppa, Artem Sokolov, Mohit Iyyer and Slav Petrov for several useful discussions during the course of this project. We are also grateful to Rajiv Teja Nagipogu, Shachi Dave, Bhuthesh R, Parth Kothari, Bhanu Teja Gullapalli and Simran Khanuja for helping us annotate model outputs in several Indian languages during pilot experiments. This work was mostly done during Kalpesh Krishna (KK)'s internship at Google Research India, hosted by Bidisha Samanta and Partha Talukdar. KK was partly supported by a Google PhD Fellowship.

\section*{Ethical Considerations}

Recent work has highlighted issues of stylistic bias in text generation systems, specifically machine translation systems~\citep{hovy2020you}. We acknowledge these issues, and consider style transfer and style-controlled generation technology as an opportunity to work towards fixing them (for instance, gender neutralization as presented in \sectionref{sec:analysis}). Note that it is important to tread down this path carefully --- In Chapter 9, ~\citet{blodgett2021sociolinguistically} argue that style is inseparable from social meaning (as originally noted by~\citealp{eckert2008variation}), and humans may perceive automatically generated text very differently compared to automatic style classifiers.

Our models were trained on 32 Google Cloud TPUs. As discussed in \appendixref{sec:model-training}, the \ur~\& \urindic~model take roughly 18 hours to train. The \diffur-* and \exemp~models are much cheaper to train (2 hours) since we finetune the pretrained \ur-* models. The Google 2020 environment report mentions,\footnote{\url{https://www.gstatic.com/gumdrop/sustainability/google-2020-environmental-report.pdf}} ``TPUs are highly efficient chips which have been specifically designed for machine learning applications''. These accelerators run on Google Cloud, which is carbon neutral today, and is aiming to ``run on carbon-free energy, 24/7, at all of Google's data centers by 2030'' (\url{https://cloud.google.com/sustainability}).

\bibliography{anthology,custom,bib/journal-full,bib/acl_latex,bib/paraphrase_emnlp2020.bib}
\bibliographystyle{acl_natbib}
\newpage
\appendix
\section*{Appendices for ``\titletext''}

\section{Model training details}
\label{sec:model-training}

 To \textbf{train} the \urindic~model, we use mC4~\citep{xue-etal-2021-mt5} for the self-supervised objectives and Samanantar~\citep{ramesh2021samanantar} for the supervised translation. For creating paraphrase data for training our \diffur~models (\sectionref{sec:diffur-model}), we again leverage Indic language side of Samanantar sentence pairs. Our models are implemented in JAX~\citep{jax2018github} using the T5X library.\footnote{\url{https://github.com/google-research/t5x}} We re-use the \ur~checkpoint from~\citet{garcia2021towards}. To train the \urindic~model, we follow the setup in~\citet{garcia2021towards} and initialize the model with mT5-XL~\citep{xue-etal-2021-mt5}, which has 3.7B parameters. We fine-tune the model for 25K steps with a batch size of 512 inputs and a learning rate of 1e-3, using the objectives in \sectionref{sec:universal-rewriter-modeling}. Training was done on 32 Google Cloud TPUs which took a total of 17.5 hours. To train the \diffur~and~\diffurindic~models, we further fine-tune \ur~and \urindic~for a total of 4K steps using the objective from \sectionref{sec:diffur-model}, taking 2 hours.

\section{More Related Work}
\label{appendix:more-related-work}

\noindent \textbf{Multilingual style transfer} is mostly unexplored in prior work: a 35 paper survey by~\citet{briakou-etal-2021-ola} found only one work in Chinese, Russian, Latvian, Estonian, French~\citep{shang2019semi,tikhonov2018sounds,korotkova2019grammatical,niu2018multi}.~\citet{briakou-etal-2021-ola} further introduced XFORMAL, the first formality transfer \emph{evaluation} dataset in French, Brazilian Portugese and Italian.\footnote{We do not use this data since it does not cover Indian languages, and due to Yahoo! L6 corpus restrictions for industry researchers (confirmed via authors correspondence).} Hindi formality has been studied in linguistics, focusing on politeness~\citep{kachru2006hindi,agnihotri2013hindi,kumar-2014-developing} and code-mixing~\citep{bali2014borrowing}. Due to its prevalence in India, English-Hindi code-mixing has seen work in language modeling~\citep{pratapa2018language, samanta2019deep} and core NLP tasks~\citep{khanuja2020gluecos}. To the best of our knowledge, we are the first to study style \emph{transfer} for Indic languages.
A few prior works build models which can \textbf{control the degree of style transfer} using a scalar input~\citep{wang2019controllable, samantahierarchical}. However, these models are style-specific and require large unpaired style corpora during training. We adopt the inference-time control method used by~\citet{garcia2021towards} and notice much better controllability after our proposed fixes in \sectionref{sec:diffur-model}.

\section{More details on the translation-specific Universal Rewriter objectives}
\label{sec:details-ur-model}

In this section we describe the details of the supervised translation objective and the style-controlled translation objective used in the Universal Rewriter model. See \sectionref{sec:universal-rewriter-modeling} for details on the exemplar-based denoising objective.

\noindent \textbf{Learning translation via direct supervision}: This objective is the standard supervised translation setup, using zero vectors for style. The output language code is prepended to the input. Consider a pair of parallel sentences $(x, y)$ in languages with codes \texttt{lx}, \texttt{ly} (prepended to the input string),
\begin{align*}
    \bar{y} &= f_\text{ur}(\texttt{ly} \oplus x, \mathbf{0}) \\
    \mathcal{L}_\text{translate} &= \mathcal{L}_\text{CE}(\bar{y}, y)
\end{align*}

The Universal Rewriter is trained on English-centric translation data from the high-resource languages in OPUS-100~\citep{zhang-etal-2020-improving}.\\

\noindent \textbf{Learning style-controlled translation}: This objective emulates "style-controlled translation" in a self-supervised manner, via backtranslation through English. Consider $x_1$ and $x_2$ to be two non-overlapping spans in mC4 in language \texttt{lx},
\begin{align*}
    x_2^{\texttt{en}} &= f_\text{ur}(\texttt{en} \oplus x_2, -f_\text{style}(x_1)) \\
    \bar{x}_2 &= f_\text{ur}(\texttt{lx} \oplus x_2^{\texttt{en}}, f_\text{style}(x_1)) \\
    \mathcal{L}_\text{BT} &= \mathcal{L}_\text{CE}(\bar{x}_2, x_2)
\end{align*}

\section{Choice of Exemplars}
\label{appendix:exemplar-choice}

\begin{figure}[h!]
    \centering
    \includegraphics[width=0.48\textwidth]{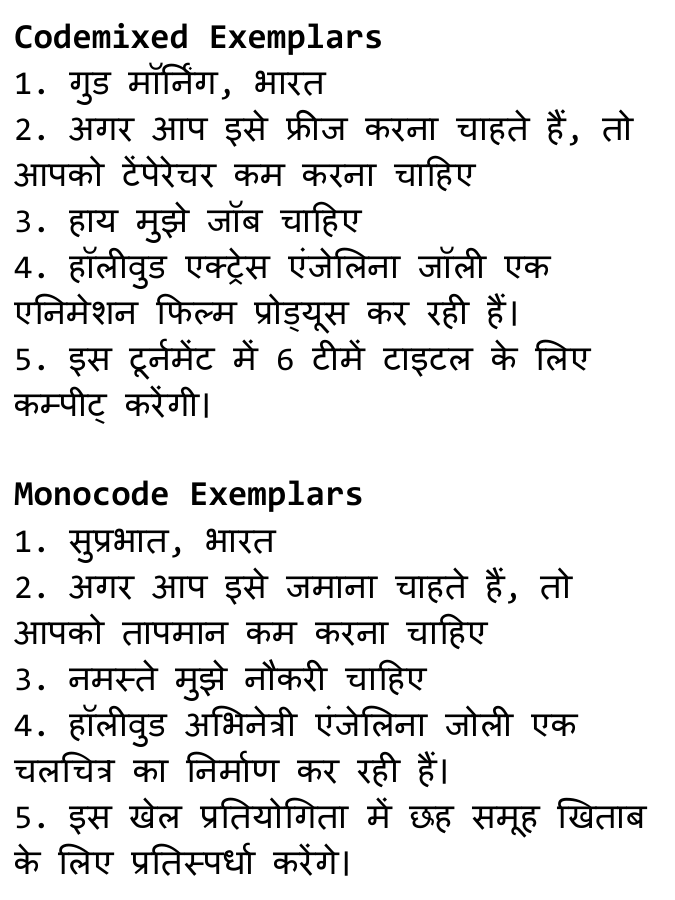}
    \caption{Exemplars used for adding code-mixing.}
    \label{fig:code-mixing-exemplars}
\end{figure}

\begin{figure}[h!]
    \centering
    \includegraphics[width=0.48\textwidth]{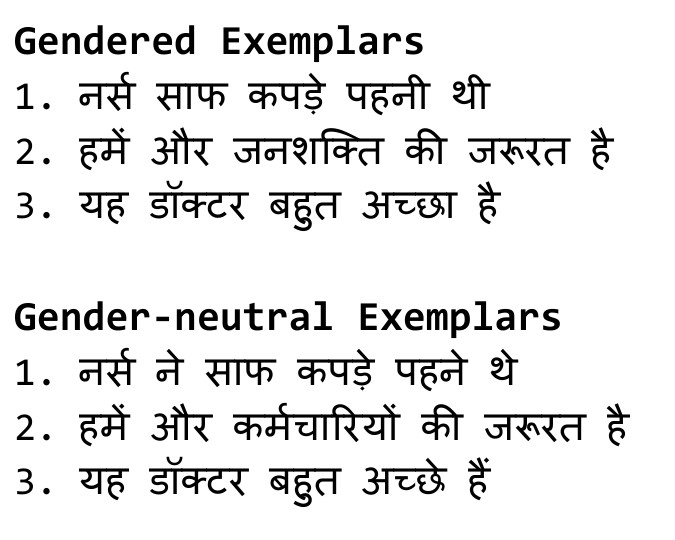}
    \caption{Exemplars used for gender neutralization.}
    \label{fig:gender-exemplars}
\end{figure}

\noindent \textbf{Formal exemplars}

\noindent 1. This was a remarkably thought-provoking read.

\noindent 2. It is certainly amongst my favorites.

\noindent 3. We humbly request your presence at our gala in the coming week.

\noindent \textbf{Informal exemplars}

\noindent 1. reading this rly makes u think

\noindent2. Its def one of my favs

\noindent 3. come swing by our bbq next week if ya can make it\\

\begin{figure}[h!]
    \centering
    \includegraphics[width=0.48\textwidth]{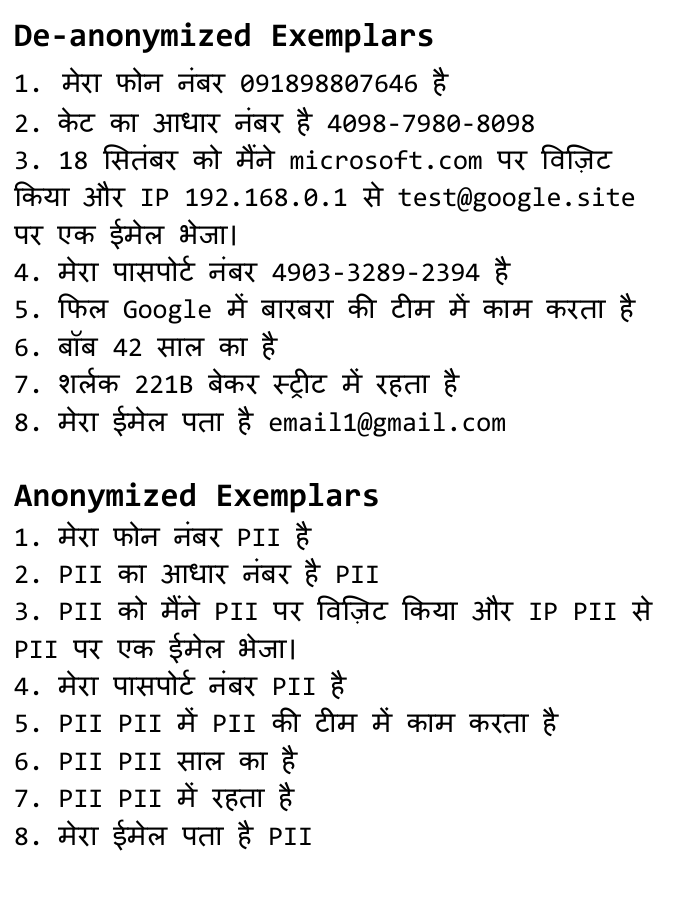}
    \caption{Exemplars used for text anonymization. All entities in the deanonymized exemplars are random.}
    \label{fig:anonymization-exemplars}
\end{figure}

\noindent \textbf{Complex exemplars}

\noindent 1. The static charges remain on an object until they either bleed off to ground or are quickly neutralized by a discharge.

\noindent 2. It is particularly famous for the cultivation of kiwifruit.

\noindent 3. Notably absent from the city are fortifications and military structures.

\noindent \textbf{Simple exemplars}

\noindent 1. Static charges last until they are grounded or discharged.

\noindent 2. This area is known for growing kiwifruit.

\noindent 3. Some things important missing from the city are protective buildings and military buildings.\\

\noindent \textbf{Positive sentiment exemplars}

\noindent 1. The most comfortable bed I've ever slept on, I highly recommend it.

\noindent 2. I loved it.

\noindent 3. The movie was fantastic.

\noindent \textbf{Negative sentiment exemplars}

\noindent 1. The most uncomfortable bed I've ever slept on, I would never recommend it.

\noindent 2. I hated it.

\noindent 3. The movie was awful.

\section{Evaluation Appendix}

\subsection{Multilingual Classifier Selection}
\label{appendix:classifier}

Due to the absence of a style classification dataset in Indic languages, we built our multilingual classifier drawing inspiration from recent research in zero-shot cross-lingual transfer~\citep{conneau-etal-2018-xnli, conneau2019cross, pfeiffer-etal-2020-mad}. We experimented with three zero-shot transfer techniques while selecting our classifiers for evaluating multilingual style transfer. \\

\noindent \textsc{translate train}: The first technique uses the hypothesis that style is preserved across translation. We classify the style of English sentences in the Samanantar translation dataset~\cite{ramesh2021samanantar} using a style classifier trained on English formality data from~\citet{krishna-etal-2020-reformulating}. We use the human translated Indic languages sentences as training data. This training data is used to fine-tune a large-scale multilingual language model. \\

\noindent \textsc{zero-shot}: The second technique fine-tunes large-scale multilingual language models on a English style transfer dataset, and applies it zero-shot on multilingual data during inference. \\

\noindent \textsc{mad-x}: Introduced by~\citet{pfeiffer-etal-2020-mad}, this technique is similar to \textsc{zero-shot} but additionally uses language-specific parameters (``adapters'') during inference. These language-specific adapters have been originally trained using masked language modeling on the desired language data. \\ 

\noindent \textbf{Dataset for evaluating classifiers}: We conduct our experiments on Hindi formality classification, leveraging our evaluation datasets from \sectionref{sec:meta-eval-dataset}. We removed pairs which did not have full agreement across the three annotators and those pairs which had the consensus rating of ``Equal'' formality. This filtering process leaves us with 316 pairs in Hindi (out of 1000). In our experiments, we check whether the classifiers give a higher score to the more formal sentence in the pair.\\

\noindent \textbf{Models}: We leverage the multilingual classifiers open-sourced\footnote{\url{https://github.com/martiansideofthemoon/style-transfer-paraphrase/blob/master/README-multilingual.md}} by~\citet{krishna-etal-2020-reformulating}. These models have been trained on the English GYAFC formality classification dataset~\citep{rao-tetreault-2018-dear}, and have been shown to be effective on the XFORMAL dataset~\citep{briakou-etal-2021-ola}  for formality classification in Italian, French and Brazilian Portuguese.\footnotemark[13] These classifiers were trained on preprocessed data which had trailing punctuation stripped and English sentences lower-cased, encouraging the models to focus on lexical and syntactic choices. As base multilingual language models, we use (1) mBERT-base from~\citet{devlin-etal-2019-bert}; (2) XLM-RoBERTa-base from~\citet{conneau-etal-2020-unsupervised}.\\

\noindent \textbf{Results}: Our results on Hindi are presented in \tableref{tab:multilingual-classifier-eval-hindi} and other languages in \tableref{tab:multilingual-classifier-eval-indic}. Consistent with~\citet{pfeiffer-etal-2020-mad}, we find \textsc{mad-x} to be a superior zero-shot cross lingual transfer method compared to baselines. We also find XLM-R has better multilingual representations than mBERT. Unfortunately, AdapterHub~\citep{pfeiffer2020AdapterHub} has XLM-R language adapters available only for Hindi \& Tamil (among Indic languages). For other languages we use the \textsc{zero-shot} technique on XLM-R, consistent with the recommendations\footnotemark[13] provided by~\citet{krishna-etal-2020-reformulating} based on their experiments on XFORMAL~\citep{briakou-etal-2021-ola}.

\begin{table}[h]
\begin{center}
\begin{tabular}{ llr } 
 \toprule
Method & Model & Accuracy ($\uparrow$) \\
\midrule
\textsc{translate train} & mBERT & 66\% \\
\textsc{zero-shot} & mBERT & 72\% \\
 & XLM-R & 76\% \\
\textsc{mad-x} & XLM-R & \textbf{81}\% \\
\bottomrule
\end{tabular}
\end{center}
\caption{Hindi formality classification accuracy on our crowdsourced dataset (\sectionref{sec:meta-eval-dataset}) using different cross-lingual transfer methods. Our results indicate that \textsc{mad-x} is the most effective method, and XLM-R is a better pretrained model than mBERT.}
\label{tab:multilingual-classifier-eval-hindi}
\end{table}

\begin{table}[h]
\begin{center}
\begin{tabular}{ lrr } 
 \toprule
Language & mBERT & XLM-R \\
\midrule
bn & 65.3\% & 82.2\% \\
kn & 76.3\% & 76.9\% \\
te & 72.6\% & 74.6\% \\
\bottomrule
\end{tabular}
\end{center}
\caption{Formality classification on our crowdsourced Bengali, Kannada and Telugu dataset (\sectionref{sec:meta-eval-dataset}) using the \textsc{zero-shot} technique described in \appendixref{appendix:classifier}. Results confirm the efficacy of the XLM-R classifier. See \tableref{tab:multilingual-classifier-eval-hindi} for Hindi results.}
\label{tab:multilingual-classifier-eval-indic}
\end{table}

\subsection{Semantic Similarity Model Selection}
\label{appendix:semantic-similarity}

We considered three models for evaluating semantic similarity between the input and output:\\

\noindent (1) LaBSE~\citep{feng2020language};

\noindent (2) m-USE~\citep{yang-etal-2020-multilingual};

\noindent (3) multilingual Sentence-BERT~\citep{reimers-gurevych-2020-making}, the knowledge-distilled variant \texttt{paraphrase-xlm-r-multilingual-v1} \\

\noindent Among these models, only LaBSE has support for all the Indic languages we were interested in. No Indic language is supported by m-USE, and multilingual Sentence-BERT has been trained on parallel data only for Hindi, Gujarati and Marathi among our Indic languages. However, in terms of Semantic Textual Similarity (STS) benchmarks~\citep{cer-etal-2017-semeval} for English, Arabic \& Spanish, m-USE and Sentence-BERT outperform LaBSE (Table 1 in~\citealp{reimers-gurevych-2020-making}).\\

\noindent \textbf{LaBSE correlates better than Sentence-BERT with our human-annotated formality dataset}: We measured the Spearman's rank correlation between the semantic similarity annotations on our human-annotated formality datasets (\sectionref{sec:meta-eval-dataset}). We discarded ~10\% sentence pairs which had no agreement among three annotators and took the majority vote for the other sentence pairs. We assigned ``Different Meaning'' a score of 0, ``Slight Difference in Meaning'' a score of 1 and ``Approximately Same Meaning'' a score of 2 before measuring Spearman's rank correlation. In \tableref{tab:semantic-similarity-correlation} we see a stronger correlation of human annotations with LaBSE compared to Sentence-BERT, especially for languages like Bengali, Kannada for which Sentence-BERT did not see parallel data. 

\begin{table}[h]
\begin{center}
\begin{tabular}{ lrrrr } 
 \toprule
Model & hi & bn & kn & te \\
\midrule
LaBSE & 0.34 & 0.49 & 0.39 & 0.25 \\
Sentence-BERT & 0.33 & 0.36 & 0.29 & 0.18  \\
\bottomrule
\end{tabular}
\end{center}
\caption{Spearman's rank correlation between different semantic similarity models and our semantic similarity human annotations collected along with formality labels. Overall, LaBSE correlates more strongly than Sentence-BERT with our annotated data.}
\label{tab:semantic-similarity-correlation}
\end{table}



\subsection{Evaluation with Different LaBSE thresholds}
\label{sec:eval-different-labse}

In \sectionref{sec:main-experiments}, we set our LaBSE threshold $L$ to 0.75. In this section, we present our evaluations with a more and less conservative value of $L$.

In \tableref{tab:formality-eval-test-multi-labse-lower}, we present results with $L=0.65$, and in \tableref{tab:formality-eval-test-multi-labse-higher} we set $L=0.85$. Compared to \tableref{tab:formality-eval-test-multi}, trends are mostly similar, with \diffur~models and \textsc{indic}~variants outperforming counterparts. Note that the absolute values of \simmetric~and \textsc{agg} metrics differ, with absolute values going down with the stricter threshold of $L=0.85$, and up with the relaxed threshold of $L=0.65$.\\

\noindent \textbf{Comparing chosen thresholds with human annotations}: To verify these three thresholds are reasonable choices, we measure the LaBSE similarity of the sentence pairs annotated by humans, and compare the LaBSE scores to human semantic similarity annotations. We pool the ``Approximately Same Meaning'' and ``Slight Difference in Meaning'' categories as ``same'', and consider only sentence pairs with a majority rating of ``same''. In \tableref{tab:semantic-similarity-labse-thresholds} we see that the chosen thresholds span the spectrum of LaBSE values for the human annotated semantically similar pairs.


\begin{table}[h]
\begin{center}
\begin{tabular}{ lrrrr } 
 \toprule
& \multicolumn{4}{c}{\% of sentence pairs > $L$} \\
Threshold $L$ & hi & bn & kn & te \\
\midrule
0.65 & 97.4 & 96.1 & 94.6 & 90.6 \\
0.75 & 83.9 & 76.1 & 68.4 & 62.6 \\
0.85 & 75.1 & 62.7 & 50.5 & 45.5 \\
\bottomrule
\end{tabular}
\end{center}
\caption{Percentage of human annotated semantically similar pairs which have a LaBSE score of at least $L$. As we increase the threshold $L$, we see this percentage substantially reduces, indicating our chosen thresholds are within the range of variation in LaBSE scores for semantically similar sentences.}
\label{tab:semantic-similarity-labse-thresholds}
\end{table}

\subsection{More Crowdsourcing Details}
\label{appendix:crowdsourcing-setup-details}

\begin{figure*}[t]
    \centering
    \includegraphics[width=0.49\textwidth]{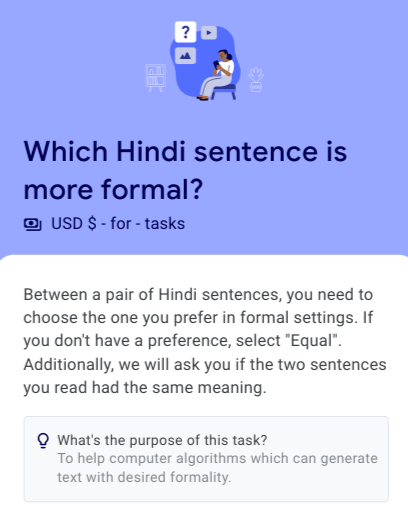}
    \includegraphics[width=0.49\textwidth]{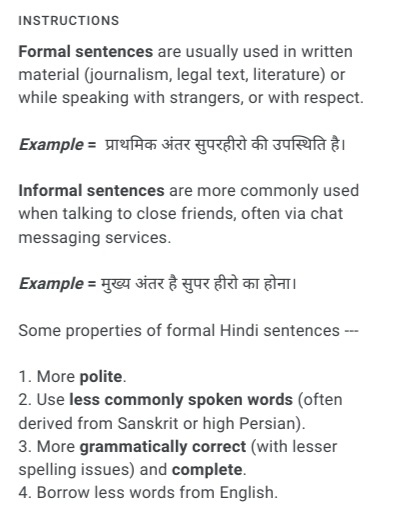}
    \\
    \includegraphics[width=0.49\textwidth]{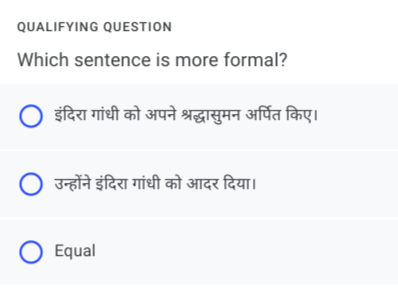}
    \includegraphics[width=0.49\textwidth]{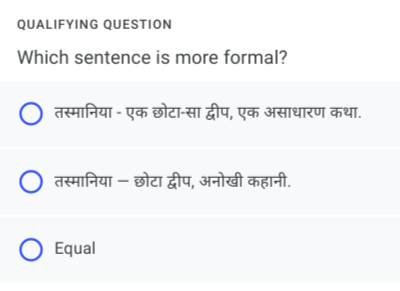}
    \\
    \includegraphics[width=0.49\textwidth]{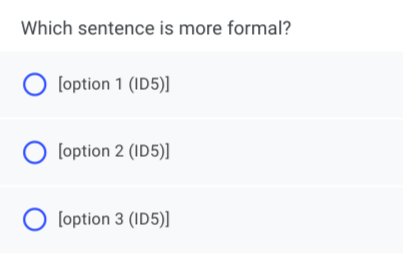}
    \includegraphics[width=0.49\textwidth]{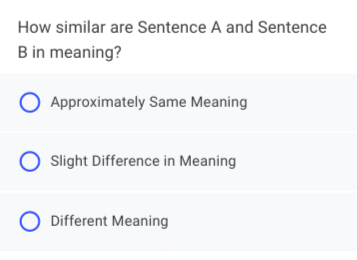}
    \caption{Our crowdsourcing interface on Task Mate, used to build our formality evaluation datasets (\sectionref{sec:meta-eval-dataset}) and conduct human evaluations (\sectionref{sec:human-evaluation}). The first row shows our landing page and instruction set derived from our conversations with professional linguists. The second row shows our qualification questions for formality classification, and the third row shows templates for the two questions asked to crowdworkers per pair.}
    \label{fig:taskmate-interface}
\end{figure*}

In \figureref{fig:taskmate-interface}, we show screenshots of our crowdsourcing interface along with all the instructions shown to crowdworkers. The instructions were written after consulting professional Indian linguists. Each crowdworker was allowed to annotate a maximum of 50 different sentence pairs per language, paying them \$0.05 per pair. For formality classification, we showed crowdworkers two sentences and asked them to choose which one is more formal. Crowdworkers were allowed to mark ties using an ``\emph{Equal}'' option. For semantic similarity annotation, we showed crowdworkers the sentence pair and provided three options --- ``\emph{approximately same meaning}'', ``\emph{slight difference in meaning}'', ``\emph{different meaning}'', to emulate a 3-point Likert scale. While performing our human evaluation (\sectionref{sec:human-evaluation}), we use a 0.5 \simmetric~score for ``\emph{slight difference in meaning}'' and a 1.0 \simmetric~score for ``\emph{approximately same meaning}'' annotations. For every system considered, we analyzed the same set of 200 input sentences for style transfer performance, and 100 of those sentences for evaluating controllability. We removed sentences which were exact copies of the input (after removing trailing punctuation) or were in the wrong language to save annotator time and cost. When outputs were exact copies of the input, we assigned \simmetric~= 100, \textsc{acc} = 0, \textsc{agg} = 0.

In \tableref{tab:kappa-crowdsource-formal} and \tableref{tab:kappa-crowdsource-sim} we show the inter-annotator agreement statistics. We measure Fleiss Kappa~\citep{fleiss1971measuring}, Randolph Kappa~\citep{randolph2005free,warrens2010inequalities}, the fraction of sentence pairs with total agreement between the three annotators and the fraction of sentence pairs with no agreement.\footnote{The $\kappa$ scores are measured using the library \url{https://github.com/statsmodels/statsmodels}.} In the table we can see all agreement statistics are well away from a uniform random annotation baseline, indicating good agreement.

\begin{table}[h]
\small
\begin{center}
\begin{tabular}{ lrrrr } 
 \toprule
 & F-$\kappa$ & R-$\kappa$ & all agree & none agree \\
 \midrule
 Random & 0.00 & 0.00 & 11.1\% & 22.2\% \\
 hi & 0.21 & 0.28 & 32.8\% & 10.2\% \\
 bn & 0.33 & 0.40 & 43.8\% & 7.2\% \\
 kn & 0.22 & 0.31 & 35.0\% & 7.7\% \\
 te & 0.21 & 0.31 & 36.0\% & 9.3\% \\
\bottomrule
\end{tabular}
\end{center}
\caption{Fleiss kappa (F-$\kappa$), Randolph kappa (R-$\kappa$), and agreement scores of crowdsourcing for \textbf{formality classification}. All $\kappa$ scores are well above a random annotation baseline, indicating fair agreement.}
\label{tab:kappa-crowdsource-formal}
\end{table}

\begin{table}[h]
\small
\begin{center}
\begin{tabular}{ lrrrr } 
 \toprule
 & F-$\kappa$ & R-$\kappa$ & all agree & none agree \\
 \midrule
  Random & 0.00 & 0.00 & 11.1\% & 22.2\% \\
 hi & 0.10 & 0.27 & 32.6\% & 11.8\% \\
 bn & 0.24 & 0.34 & 38.7\% & 10.2\% \\
 kn & 0.13 & 0.25 & 30.8\% & 11.3\% \\
 te & 0.10 & 0.31 & 36.1\% & 9.7\% \\
\bottomrule
\end{tabular}
\end{center}
\caption{Fleiss kappa (F-$\kappa$), Randolph kappa (R-$\kappa$), and agreement scores of crowdsourcing for \textbf{semantic similarity}. All $\kappa$ scores are well above a random annotation baseline, indicating fair agreement.}
\label{tab:kappa-crowdsource-sim}
\end{table}

\subsection{Fluency Evaluation}
\label{appendix:no-fluency-evaluation}

Unlike some prior works, we \textbf{avoid evaluation of output fluency} due to the following reasons: (1) lack of fluency evaluation tools for Indic languages;\footnote{A potential tool for fluency evaluation in future work is LAMBRE~\citep{pratapa-etal-2021-evaluating}. However, the original paper does not evaluate performance on Indic languages and the grammars for Indic languages would need to collected / built.} (2) fluency evaluation often discriminates against styles which are out-of-distribution for the fluency classifier, as discussed in Appendix A.8 of~\citet{krishna-etal-2020-reformulating}; (3) several prior works~\citep{pang-2019-towards,mir-etal-2019-evaluating,krishna-etal-2020-reformulating} have recommended against using perplexity of style language models for fluency evaluation since it is unbounded and favours unnatural sentences with common words; (4) large language models are known to produce fluent text as perceived by humans~\citep{ippolito-etal-2020-automatic,akoury-etal-2020-storium}, reducing the need for this evaluation.

\subsection{Details of other individual metrics}
\label{appendix:other-individual-details}

\noindent \textbf{Language Consistency (\textsc{lang})}: Since our semantic similarity metric LaBSE is language-agnostic, it tends to ignore accidental translations, which are common errors in large multilingual transformers~\citep{xue2021byt5, xue-etal-2021-mt5}, especially the Universal Rewriter (\sectionref{sec:universal-rewriter-shortcomings}). Hence, we check whether the output sentence is in the same language as the input, using \texttt{langdetect}.\footnote{This \href{https://github.com/Mimino666/langdetect}{package} is the Python port of~\citet{nakatani2010langdetect}.}\\

\noindent \textbf{Output Diversity (\copymetric, \fmetric)}: As discussed in \sectionref{sec:universal-rewriter-shortcomings}, the Universal Rewriter has a strong tendency to copy the input verbatim. We build two metrics to measure output diversity compared to the input, which have been previously used for extractive question answering evaluation~\citep{rajpurkar-etal-2016-squad}. The first metric \copymetric~measures the fraction of outputs which were copied verbatim from the input. This is done after removing trailing punctuation, to penalize models generations which solely modify punctuation. A second metric \fmetric~measures the unigram overlap F1 score between the input and output. A diverse style transfer system should minimize both \copymetric~and \fmetric.

\section{More Controllability Evaluations}
\label{appendix:more-control-evals}

We follow the setup in \sectionref{sec:evaluating-control} to first compute a $\lambda_\text{max}$ per system. We then compute the following,\\

\noindent 1. \textbf{Style Transfer Performance} (\relagg): An ideal system should have good overall performance (\sectionref{sec:aggregation-overall-style-transfer}) across different values in the range $\Lambda$.

\noindent 2. \textbf{Average Style Score Increase} (\stylechange): As our control value increases, we want the classifier's target style score (compared to the input) to increase. Additionally, we want the style score increase of $\lambdamax$ to be as high as possible, indicating the system can span the range of classifier scores.

\noindent 3. \textbf{Style Calibration to $\lambda$} (\stylecalib, \stylecalibinp): As defined in \sectionref{sec:evaluating-control}.  We additionally also measure calibration by including the input sentence $x$ in the \stylecalib($x$) calculation, treating it as the output for $\lambda = 0$ (no style transfer). Here, calibration is averaged over a total of $n=6$ $(\lambda_1, \lambda_2)$ pairs. We call this metric \stylecalibinp.

\noindent A detailed breakdown of performance by different metrics for every model is shown in \tableref{tab:hindi-formality-eval-test-control-full}.

\section{Ablation Studies}
\label{appendix:ablation-studies}

\subsection{Ablation Study for \diffur~design}
\label{appendix:ablation-explanations}

This section describes the ablation experiments conducted for the \diffur~modeling choices in \sectionref{sec:diffur-model}. We ablate a \diffurindic~model trained on Hindi paraphrase data only, and present results for Hindi formality transfer in \tableref{tab:hindi-formality-ablations}.\\

\noindent - \textbf{no paraphrase}: We replaced the paraphrase noise function with the random token dropping / replacing noise used in the denoising objective of \ur~model (\sectionref{sec:universal-rewriter-modeling}), and continued to use vector differences. As seen in \tableref{tab:hindi-formality-ablations}, this significantly increases the copy rate, which lowers the style transfer performance.\\

\noindent - \textbf{no paraphrase semantic filtering}: We keep a setup identical to \sectionref{sec:diffur-model}, but avoid the LaBSE filtering done (discarding pairs having a LaBSE score outside [0.7, 0.98]) to remove noisy paraphrases or exact copies. As seen in \tableref{tab:hindi-formality-ablations}, this decreases the semantic similarity score of the generations, lowering the overall performance.\\

\noindent - \textbf{no vector differences}: Instead of using vector differences for \diffurindic, we simply set $\mathbf{s}_\text{diff} = f_\text{style}(x)$, or the style of the target sentence. In \tableref{tab:hindi-formality-ablations}, we see this significantly decreases \simmetric~scores, and \lang~scores for $\lambda=2.0$. We hypothesize that this training encourages the model to rely more heavily on the style vectors, ignoring the paraphrase input. This could happen since the style vectors are solely constructed from the output sentence itself, and semantic information / confounding style is not subtracted out. In other words, the model is behaving more like an autoencoder (through the style vector) instead of a denoising autoencoder with stylistic supervision.\\

\noindent - \textbf{mC4 instead of Samanantar}: Instead of creating pseudo-parallel data with Samanantar, we leverage the mC4 dataset itself which was used to train the \ur~model. We backtranslate spans of text from the Hindi split of mC4 on-the-fly using the \ur~translation capabilities, and use it as the ``paraphrase noise function''. To ensure translation performance does not deteriorate during training, 50\% minibatches are supervised translation between Hindi and English. In \tableref{tab:hindi-formality-ablations}, we see decent overall performance, but the \lang~score is 6\% lower than \diffurindic. Qualitatively we found that the model often translates a few Hindi words to English while making text informal. Due to sparsity of English tokens, it often escapes penalization from \textsc{lang}.

\noindent - \textbf{mC4 + exemplar instead of target}: This setting is similar to the previous one, but in addition to the mC4 dataset we utilize the vector difference between the style vector of the exemplar span (instead of target span), and the ``paraphrase noised'' input. Results in \tableref{tab:hindi-formality-ablations} show this method is not effective, and it's important for the vector difference to model the precise transformation needed.

\subsection{Choice of Decoding Scheme}

We experiment with five decoding schemes on the Hindi formality validation set --- beam search with beam size 1, 4 and top-$p$ sampling~\citep{holtzman2020curious} with $p=0.6, 0.75, 0.9$.

\begin{figure*}[t!]
    \centering
    \includegraphics[width=0.42\textwidth]{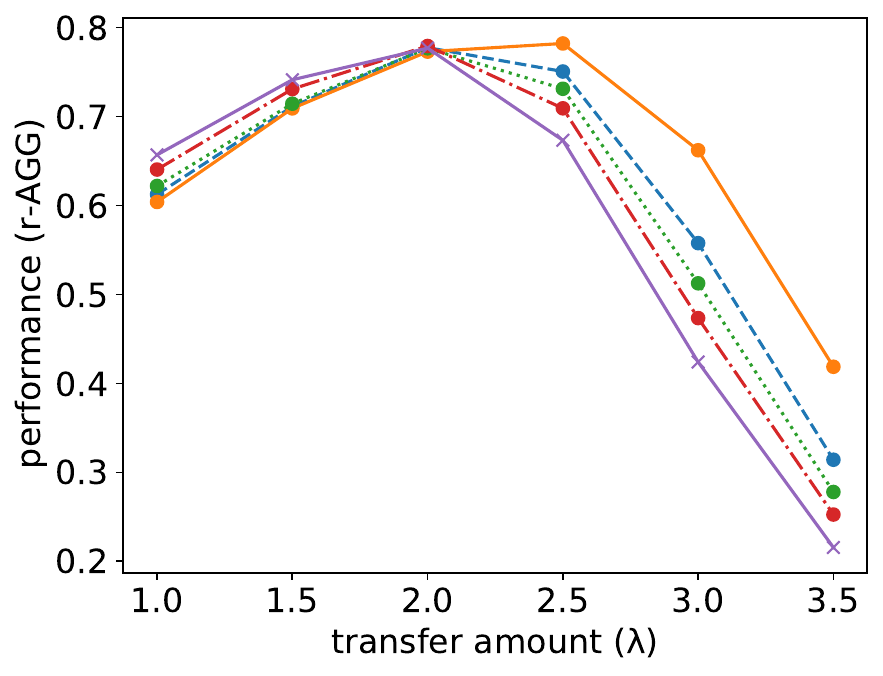}
    \includegraphics[width=0.57\textwidth]{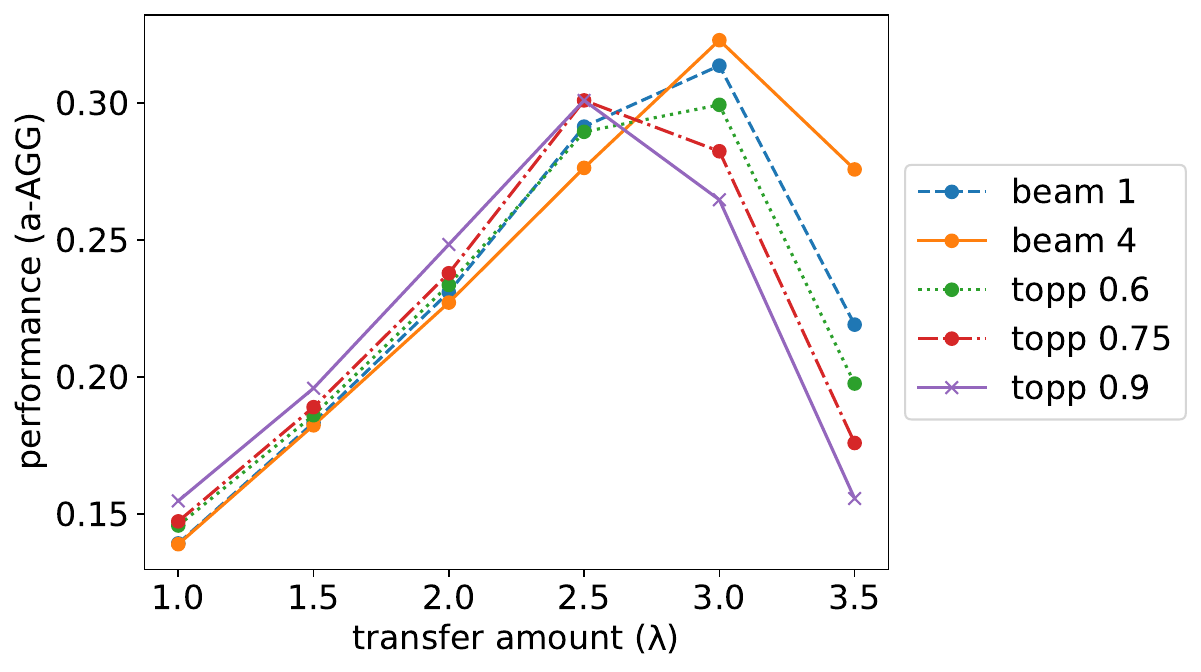}
    \caption{Variation in Hindi formality transfer (validation set) performance vs $\lambda$ with change in decoding scheme, for the \exemp~model. The plots show overall style transfer performance, using the \relagg~(left) and \absagg~(right) metrics from \sectionref{sec:aggregation-overall-style-transfer}. Beam search with beam size 4 performs best, see \tableref{tab:hindi-formality-eval-decoding-scheme} for an individual metric breakdown while keeping $\lambda = 3.0$.}
    \label{fig:ablation-decoding-graphs}
\end{figure*}

In \tableref{tab:hindi-formality-eval-decoding-scheme}, we present results at a constant style transfer magnitude ($\lambda = 3.0$). Consistent with~\citet{krishna-etal-2020-reformulating}, we find that top-$p$ decoding usually gets higher style accuracy (\relacc, \absacc) and output diversity (\fmetric, \copymetric) scores, but lower semantic similarity (\simmetric) scores. Overall beam search triumphs since the loss in semantic similarity leads to a worse performing model. In \figureref{fig:ablation-decoding-graphs}, we see a consistent trend across different magnitudes of style transfer ($\lambda$). In all our main experiments, we use beam search with beam size 4 to obtain our generations.

\subsection{Number of Training Steps}

\begin{figure*}[t]
    \centering
    \includegraphics[width=0.42\textwidth]{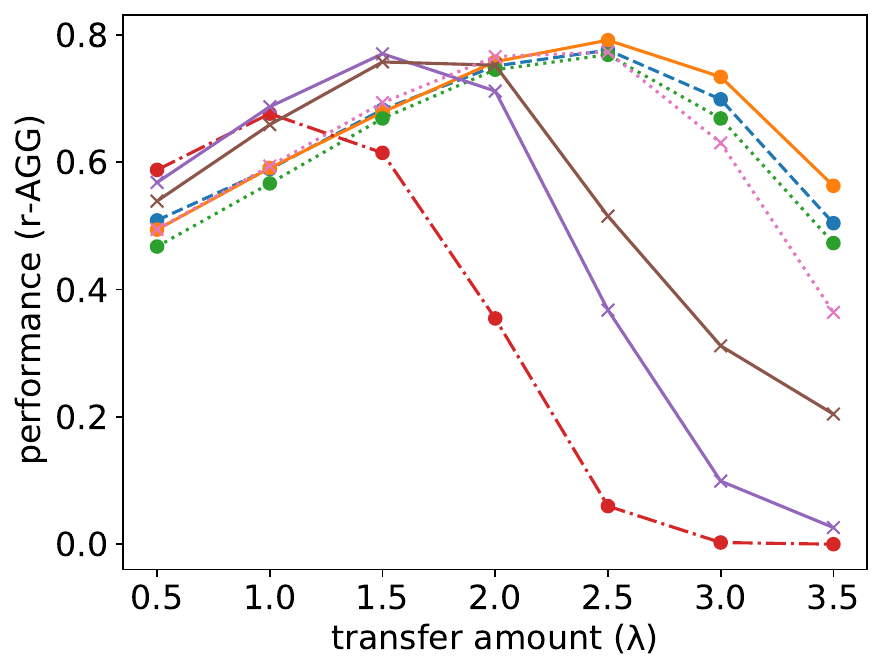}
    \includegraphics[width=0.57\textwidth]{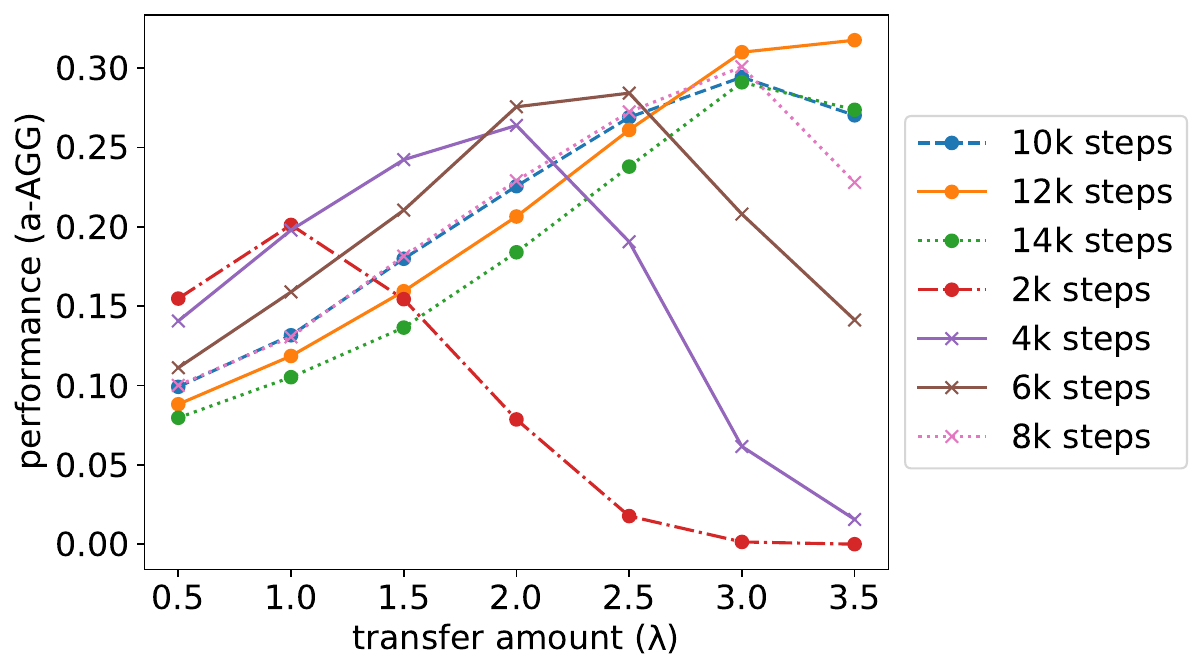}
    \includegraphics[width=0.7\textwidth]{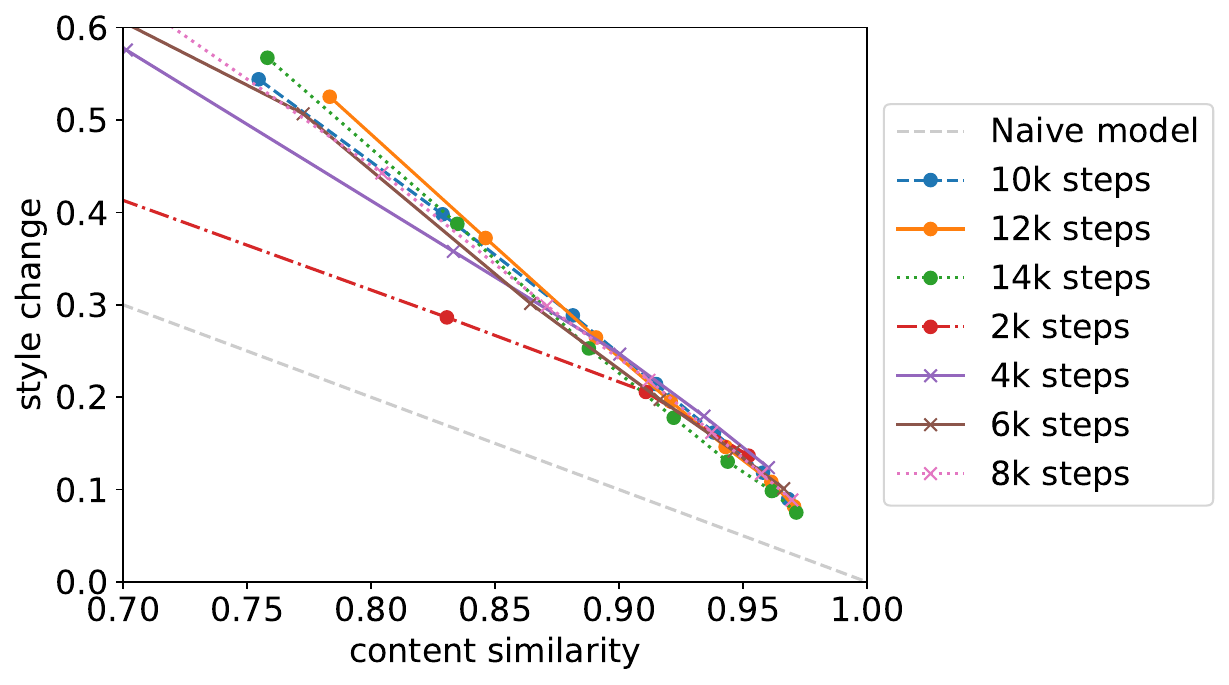}
    \caption{Variation in Hindi formality transfer validation set performance with change in number of training steps for the \exemp~model. The plots show overall style transfer performance, using the \relagg~(top-left) and \absagg~(top-right) metrics from \sectionref{sec:aggregation-overall-style-transfer}. With more training steps performance seems to improve and the peak of the graph shifts towards the right (a preference towards higher scale values). We also see more training steps leads to better controllability (bottom plot, closer to Y-axis is better), but only marginal gains after 6k steps.}
    \label{fig:ablation-steps}
\end{figure*}

In \figureref{fig:ablation-steps}, we present the variation in style transfer performance with number of training steps for our best model, the \exemp~model. We find that with more training steps performance generally improves, but improvements saturate after 8k steps. We also see the peak of the graphs (best style transfer performance) shift rightwards, indicating a preference for higher $\lambda$ values.

\section{Analysis Experiments}
\label{appendix:analysis}

\subsection{Style vectors from $f_\text{style}$ as style classifiers}
\label{appendix:style-analysis-informal-vectors}

The Universal Rewriter models succeed in learning an effective style space, useful for few-shot style transfer. But can this metric space also act as a style classifier? To explore this, we measure the cosine distance between the mean style vector of our informal exemplars,\footnote{See \appendixref{appendix:exemplar-choice} for the exemplar sentences. We found the informal exemplars more effective than formal exemplars for style classification; \appendixref{appendix:style-analysis-formal-vectors} has a comparison.} and the style vectors derived by passing human-annotated formal/informal pairs (from our dataset of \sectionref{sec:meta-eval-dataset}) through $f_\text{style}$. We only consider pairs which had complete agreement among annotators. In \tableref{tab:style-vector-analysis-informal-vecs} we see good agreement (68.2\%-80.7\%) between human annotations and the classifier derived from the metric space of the \urindic~model. Agreement is lower (67.0\%-74.3\%) for the \diffurindic~model, likely due to the stop gradient used in \sectionref{sec:diffur-model}. With \exemp, agreement jumps back up to 75\%-81.7\% since gradients flow into the style extractor as well.

\begin{table}[t!]
\begin{center}
\begin{tabular}{ lrrrr } 
 \toprule
 Model & hi & bn & kn & te\\
 \midrule
 \ur & 79.1 & 69.7 & 66.2 & 67.1  \\
 \urindic & 80.7 & 74.3 & 68.2 & 72.2 \\
\diffurindic  & 68.0 & 73.8 & 67.0 & 70.4 \\
 \exemp & 75.0 & 81.7 & 79.8 & 79.0 \\
\bottomrule
\end{tabular}
\end{center}
\caption{style vector as a classifier, measuring the cosine similarity with informal exemplar vectors.}
\label{tab:style-vector-analysis-informal-vecs}
\end{table}

\subsection{Style Vector Analysis with Formal Exemplars Vectors}
\label{appendix:style-analysis-formal-vectors}

In \appendixref{appendix:style-analysis-informal-vectors}, we saw that the metric vector space derived from the style encoder $f_\text{style}$ of various models is an effective style classifier, using the \emph{informal} exemplar vectors. In \tableref{tab:style-vector-analysis-formal-vecs}, we present a corresponding analysis using \emph{formal} exemplar vectors. Most accuracy scores are close to 50\%, implying this setup is not a very effective style classifier.

\begin{table}[h]
\begin{center}
\begin{tabular}{ lrrrr } 
 \toprule
 Model & hi & bn & kn & te \\
 \midrule
 \ur & 56.6 & 60.0 & 61.6 & 57.6 \\
 \urindic & 59.5 & 60.6 & 52.6 & 44.8 \\
 \diffurindic & 58.5 & 58.3 & 59.5 & 49.7 \\
 \exemp & 64.9 & 52.3 & 47.1 & 41.8 \\
\bottomrule
\end{tabular}
\end{center}
\caption{style vector as a classifier, measuring the cosine similarity with formal exemplar vectors.}
\label{tab:style-vector-analysis-formal-vecs}
\end{table}

\section{Full Breakdown of Results}
\label{appendix:full-results-breakdown}

\begin{figure*}[t]
    \centering
    \includegraphics[width=0.395\textwidth]{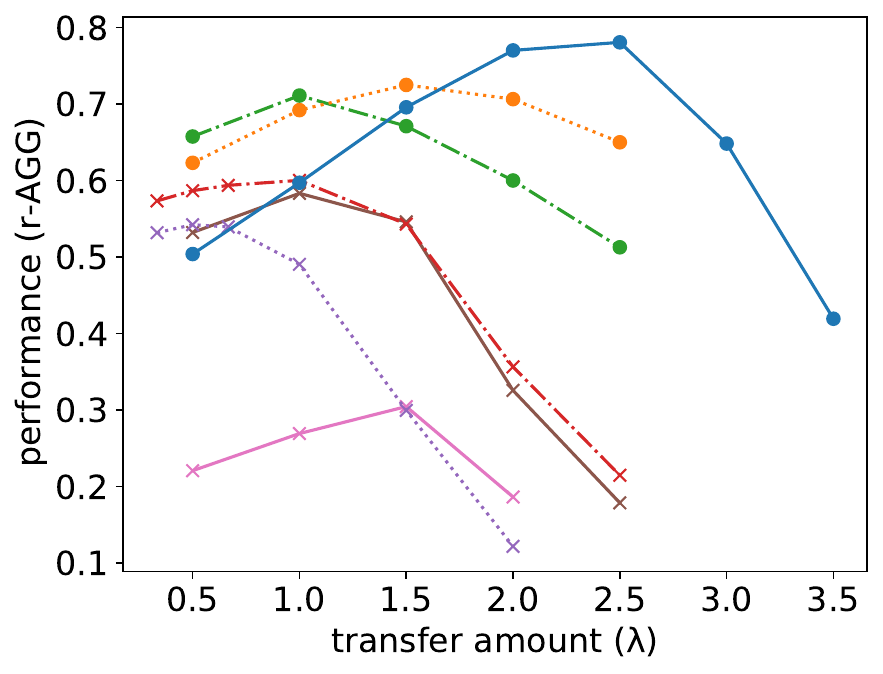}
    \includegraphics[width=0.595\textwidth]{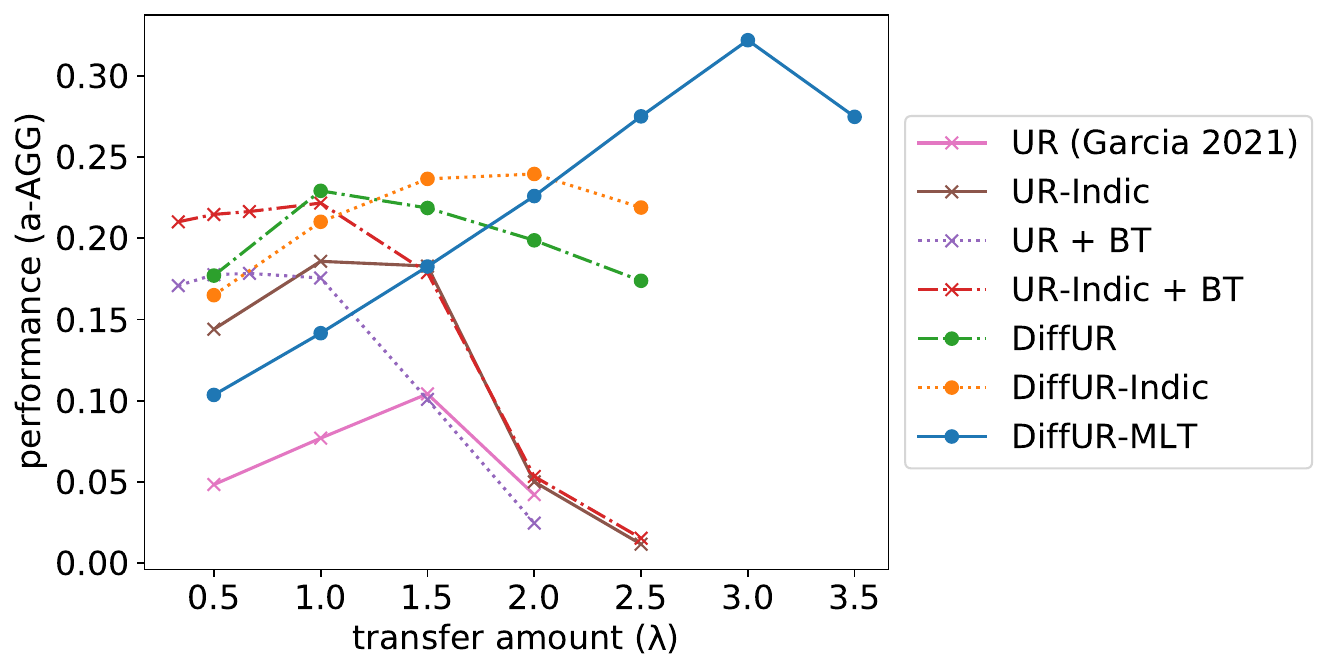}
    \includegraphics[width=0.7\textwidth]{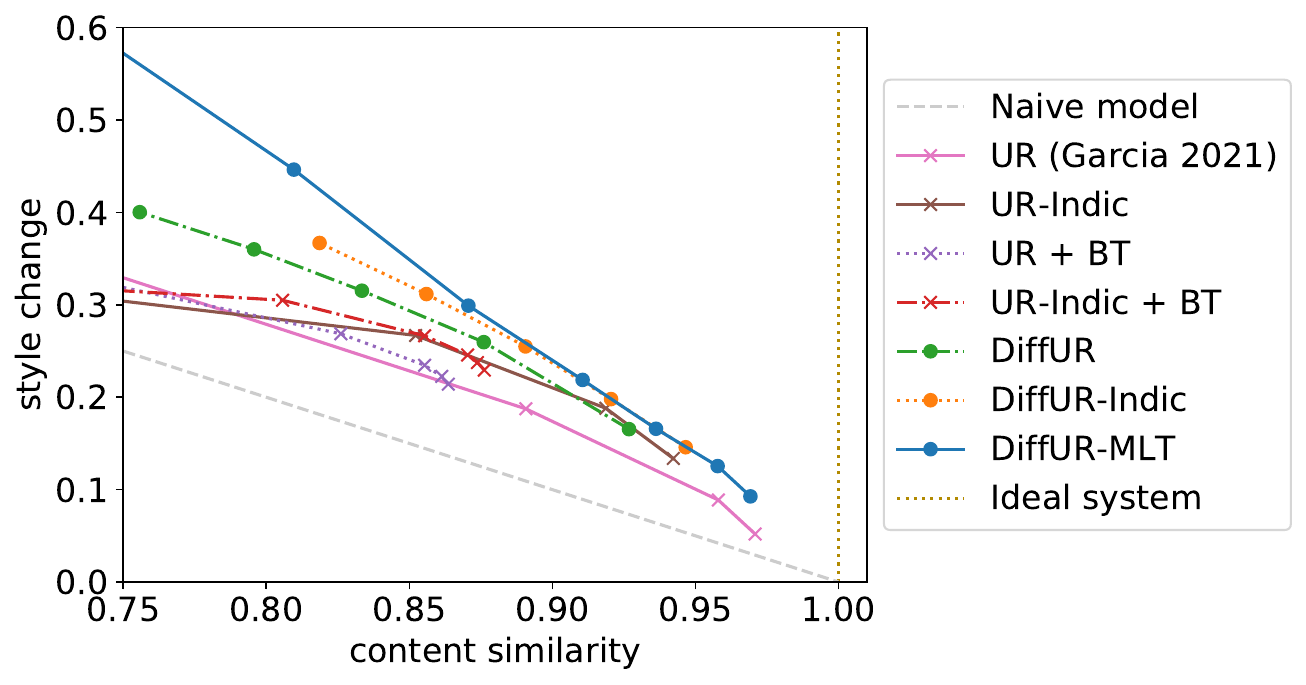}
    \caption{Variation in \textbf{Hindi} formality transfer test set performance \& control for \textbf{different models} (see \tableref{tab:hindi-test-formality-eval} for a individual metric breakdown of the models at the best performing $\lambda$). The plots show overall style transfer performance, using the \relagg~(top-left) and \absagg~(top-right) metrics from \sectionref{sec:aggregation-overall-style-transfer}. We see the \diffur~models outperform other systems across the $\lambda$ range, and get best performance with the \exemp~variant. We also see that \diffur~models, especially with \exemp, lead to better style transfer control (bottom plot, closer to $x=1$ is better), giving large style variation with $\lambda$ without loss in semantics (X-axis). }
    \label{fig:hindi-test-formality-eval-plots}
\end{figure*}

\begin{figure*}[t]
    \centering
    \includegraphics[width=0.395\textwidth]{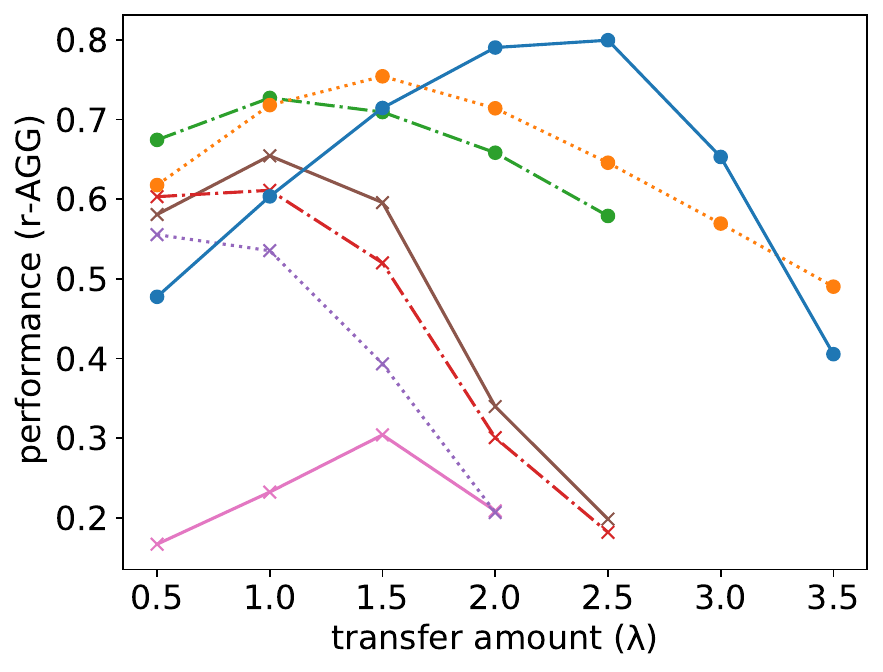}
    \includegraphics[width=0.595\textwidth]{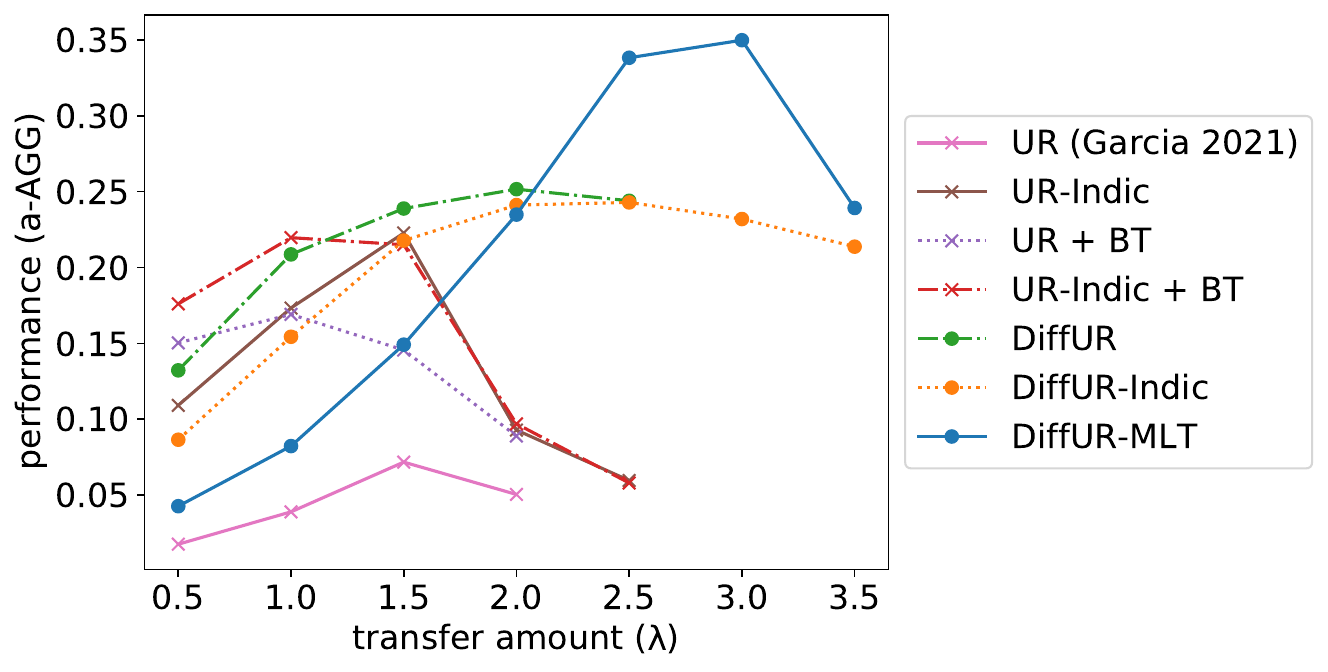}
    \includegraphics[width=0.7\textwidth]{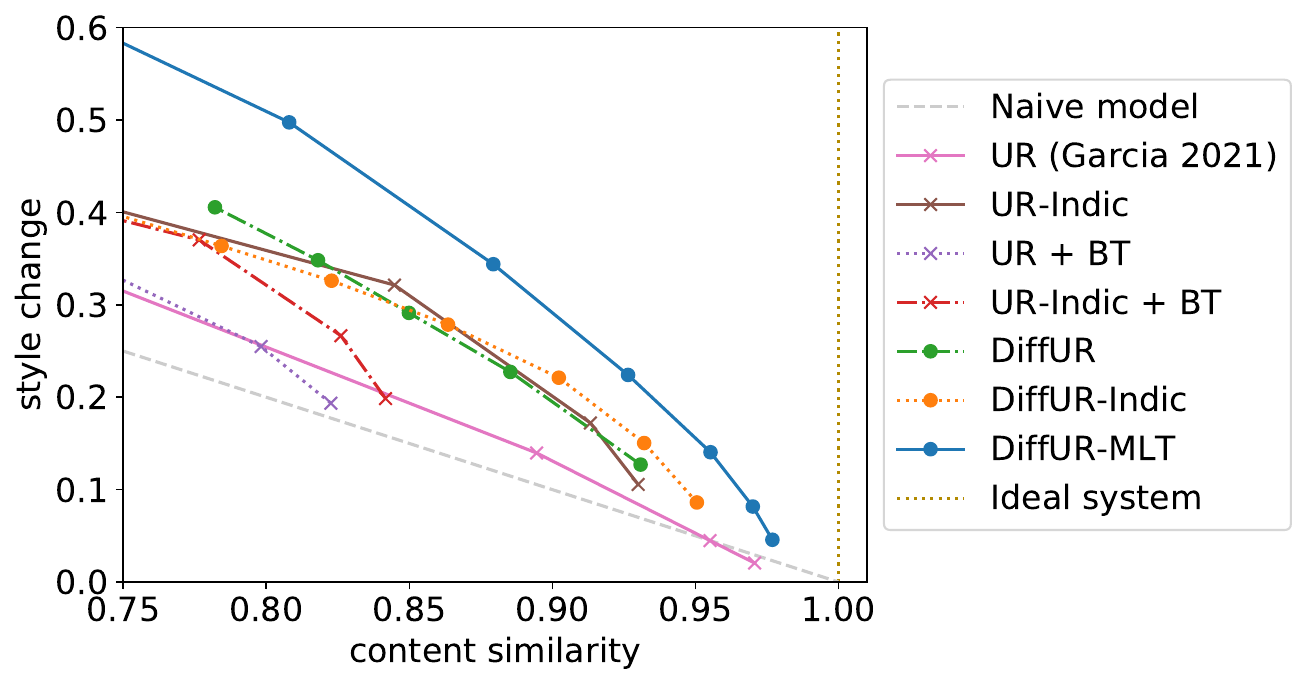}
    \caption{Variation in Bengali formality transfer test set performance \& control for different models (see \tableref{tab:bengali-test-formality-eval} for a individual metric breakdown of the models at the best performing $\lambda$). The plots show overall style transfer performance, using the \relagg~(top-left) and \absagg~(top-right) metrics from \sectionref{sec:aggregation-overall-style-transfer}. We see the \diffur~models outperform other systems across the $\lambda$ range, and get best performance with the \exemp~variant. We also see that \diffur~models, especially with \exemp, lead to better style transfer control (bottom plot, closer to $x=1$ is better), giving large style variation with $\lambda$ without loss in semantics (X-axis). }
    \label{fig:bengali-test-formality-eval-plots}
\end{figure*}

\begin{figure*}[t]
    \centering
    \includegraphics[width=0.395\textwidth]{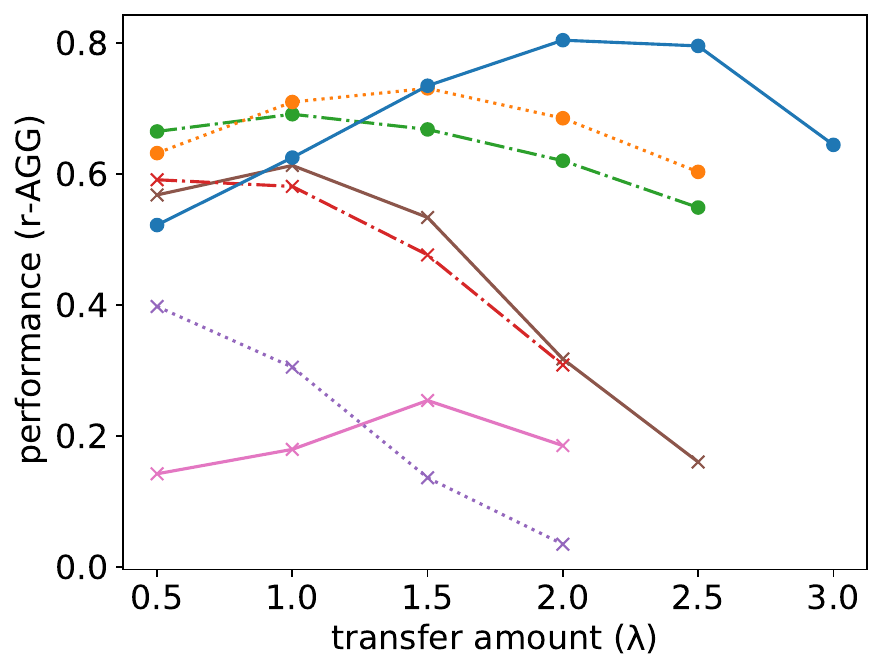}
    \includegraphics[width=0.595\textwidth]{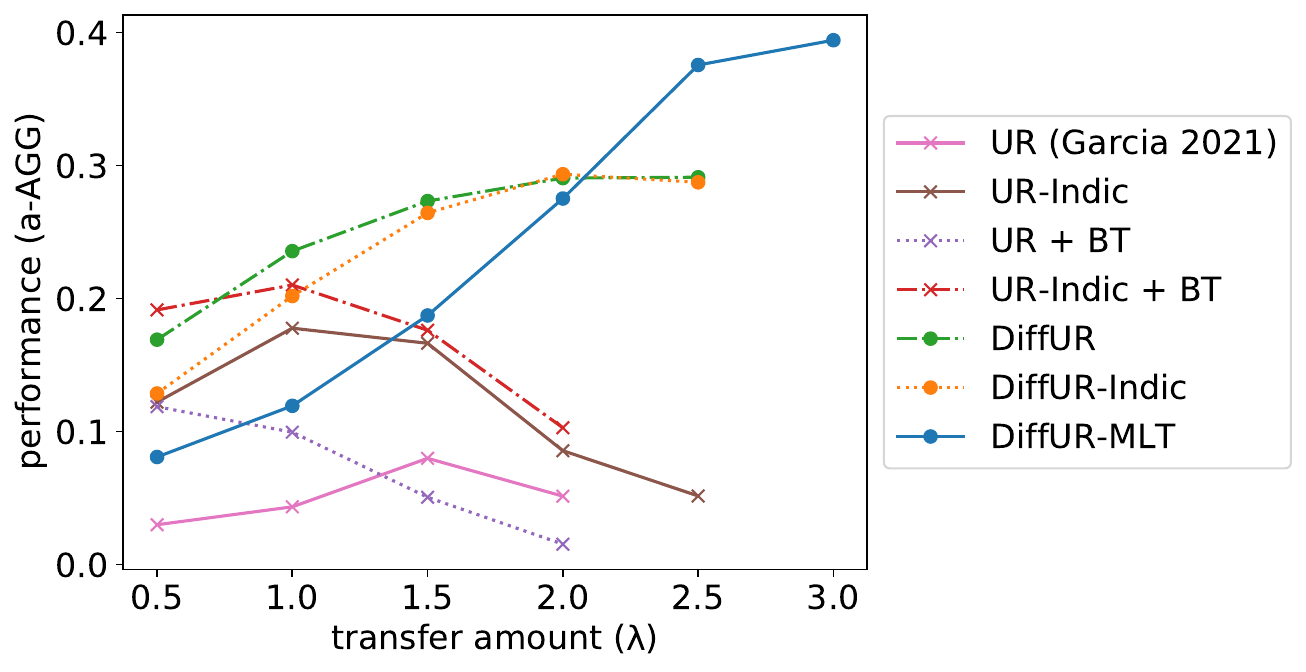}
    \includegraphics[width=0.7\textwidth]{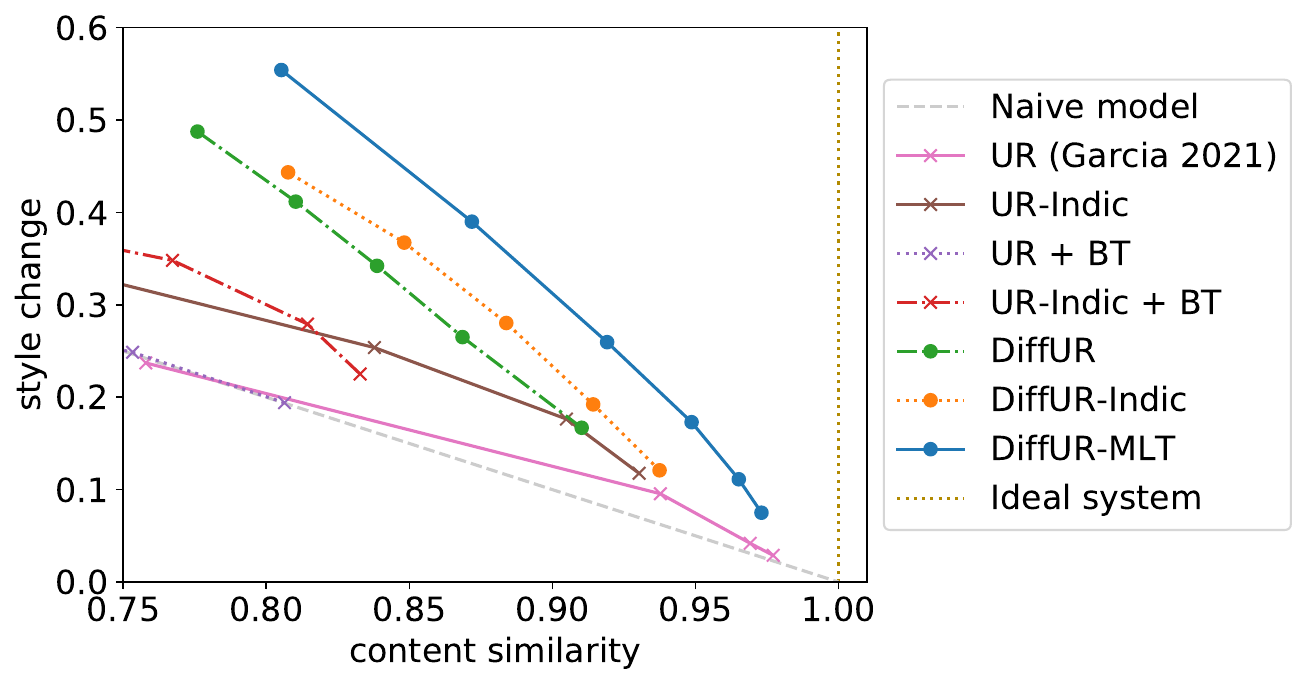}
    \caption{Variation in Kannada formality transfer test set performance \& control for different models (see \tableref{tab:kannada-test-formality-eval} for a individual metric breakdown of the models at the best performing $\lambda$). The plots show overall style transfer performance, using the \relagg~(top-left) and \absagg~(top-right) metrics from \sectionref{sec:aggregation-overall-style-transfer}. We see the \diffur~models outperform other systems across the $\lambda$ range, and get best performance with the \exemp~variant. We also see that \diffur~models, especially with \exemp, lead to better style transfer control (bottom plot, closer to $x=1$ is better), giving large style variation with $\lambda$ without loss in semantics (X-axis). }
    \label{fig:kannada-test-formality-eval-plots}
\end{figure*}

\begin{figure*}[t]
    \centering
    \includegraphics[width=0.395\textwidth]{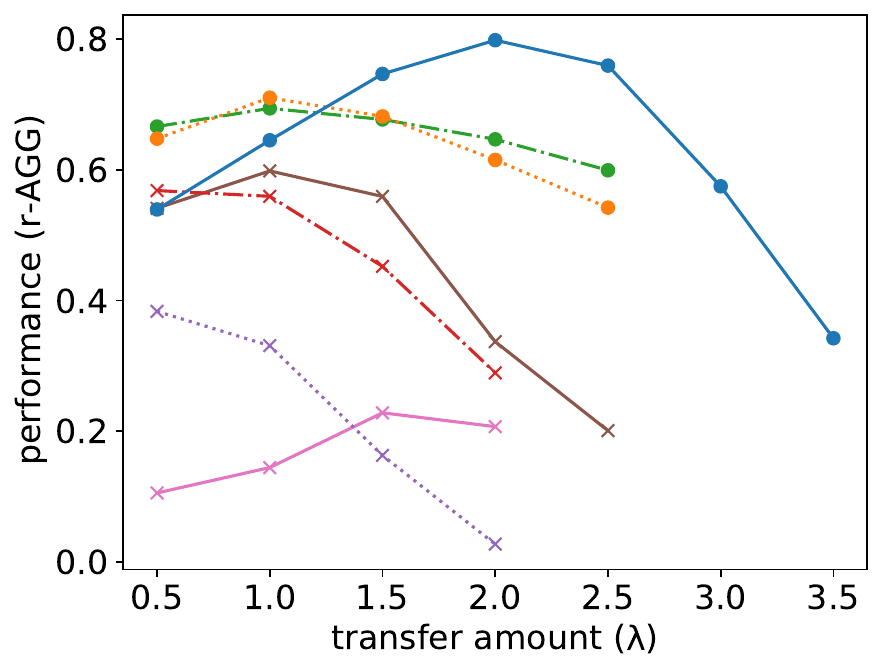}
    \includegraphics[width=0.595\textwidth]{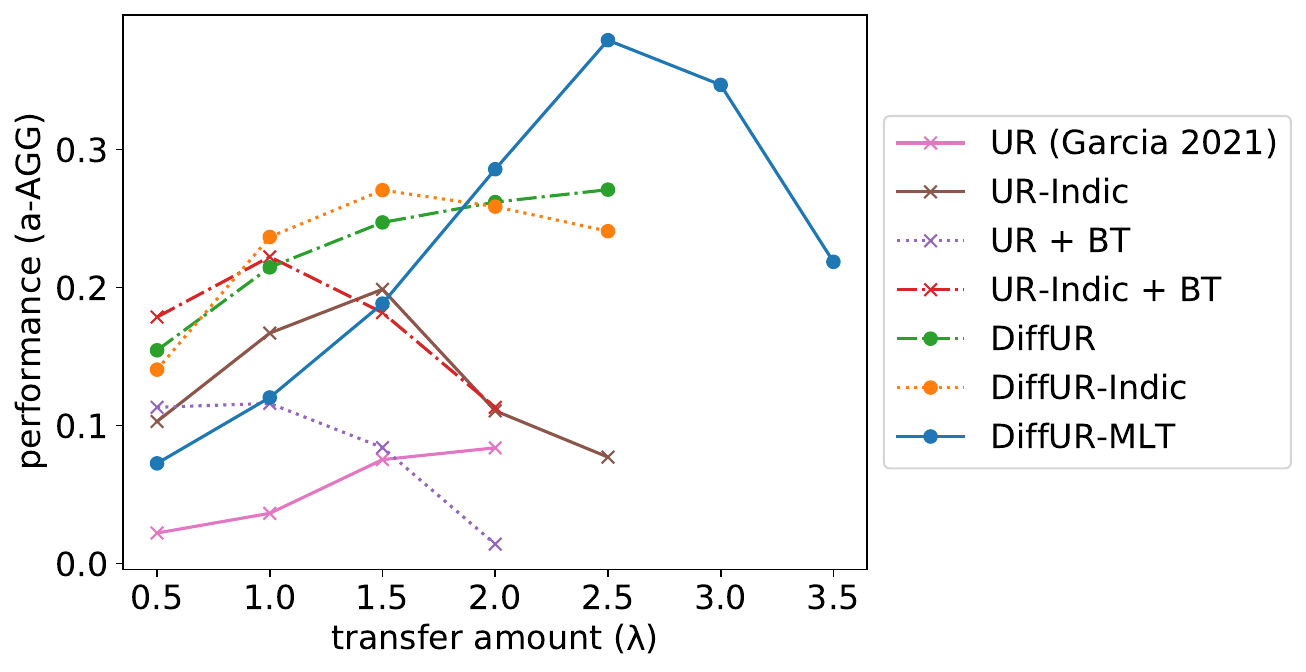}
    \includegraphics[width=0.7\textwidth]{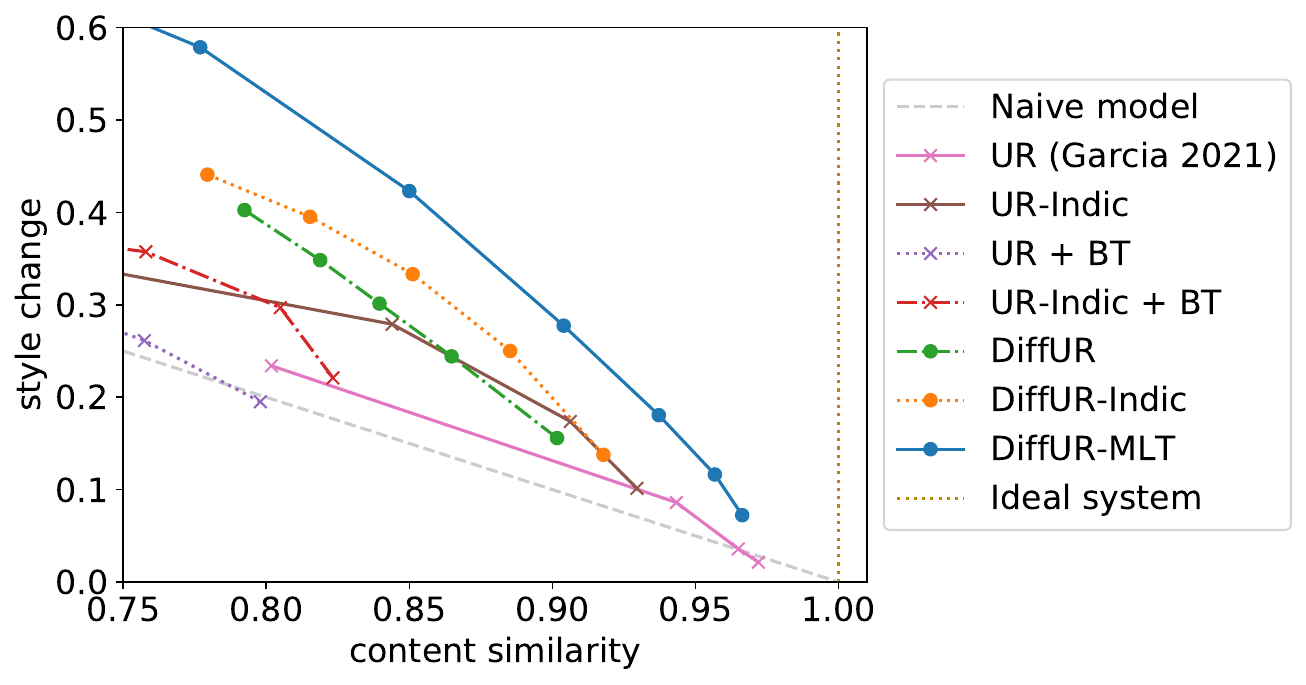}
    \caption{Variation in Telugu formality transfer test set performance \& control for different models (see \tableref{tab:telugu-test-formality-eval} for a individual metric breakdown of the models at the best performing $\lambda$). The plots show overall style transfer performance, using the \relagg~(top-left) and \absagg~(top-right) metrics from \sectionref{sec:aggregation-overall-style-transfer}. We see the \diffur~models outperform other systems across the $\lambda$ range, and get best performance with the \exemp~variant. We also see that \diffur~models, especially with \exemp, lead to better style transfer control (bottom plot, closer to $x=1$ is better), giving large style variation with $\lambda$ without loss in semantics (X-axis). }
    \label{fig:telugu-test-formality-eval-plots}
\end{figure*}

\begin{figure*}[t]
    \centering
    \includegraphics[width=0.395\textwidth]{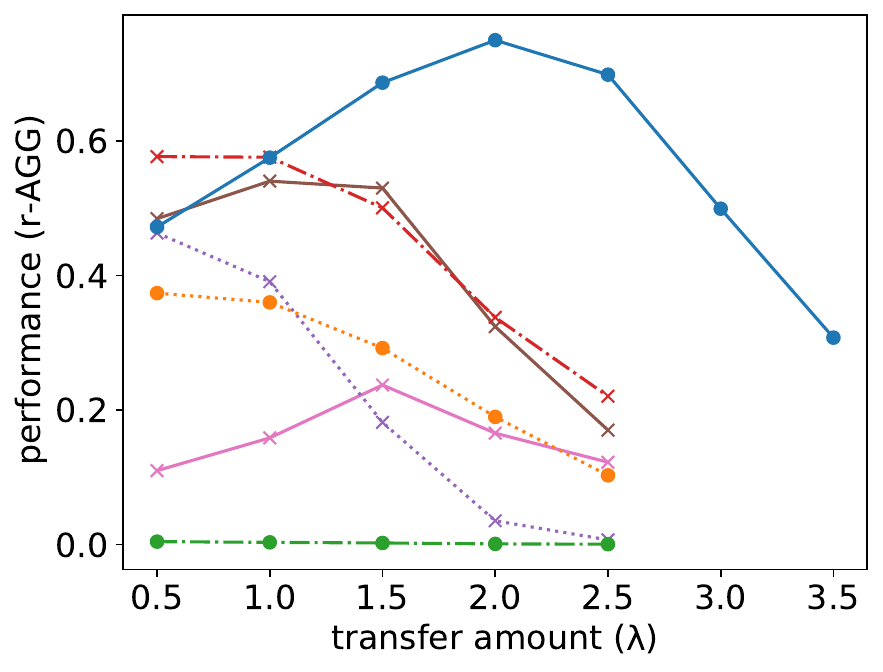}
    \includegraphics[width=0.595\textwidth]{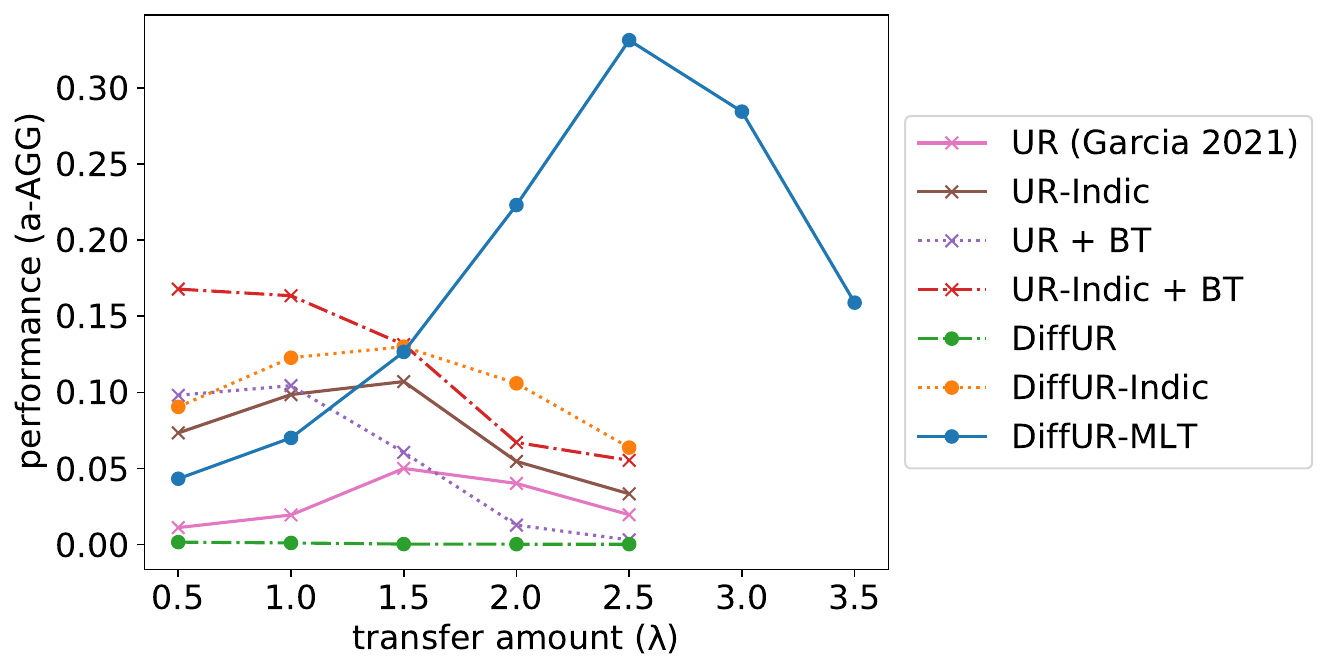}
    \includegraphics[width=0.7\textwidth]{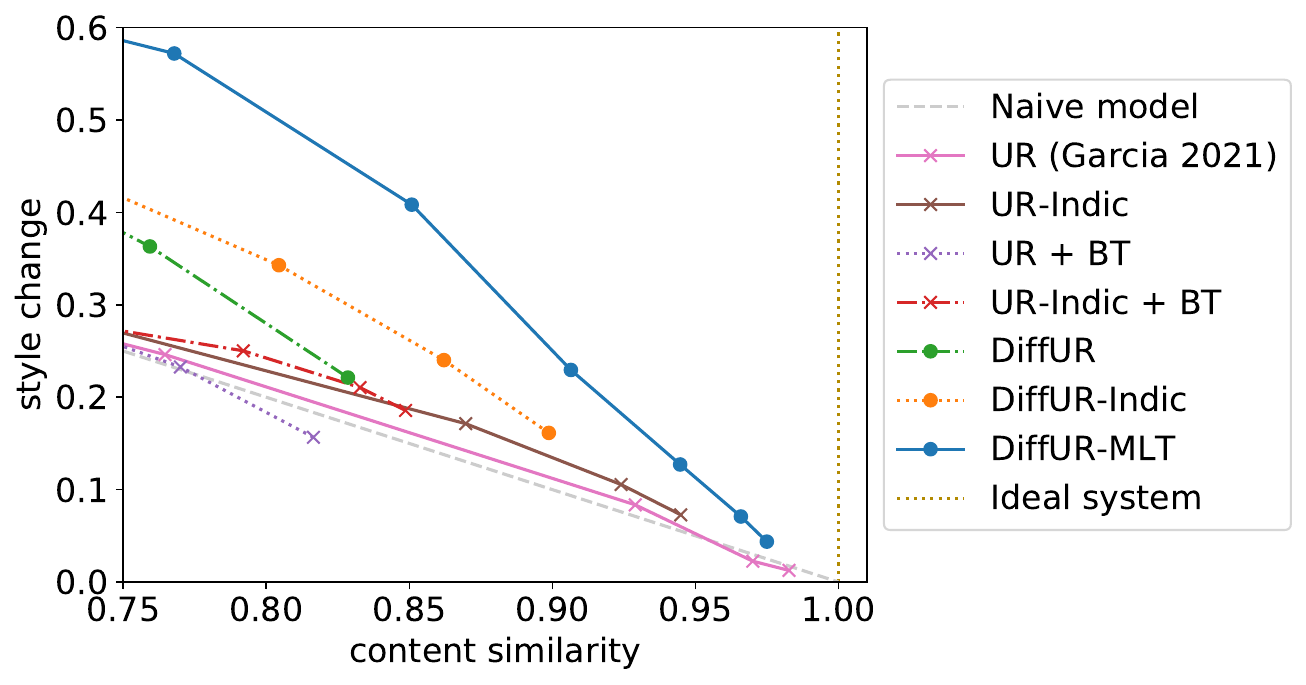}
    \caption{Variation in Gujarati formality transfer test set performance \& control for different models (see \tableref{tab:gujarati-test-formality-eval} for a individual metric breakdown of the models at the best performing $\lambda$). The plots show overall style transfer performance, using the \relagg~(top-left) and \absagg~(top-right) metrics from \sectionref{sec:aggregation-overall-style-transfer}. Note that Gujarati is a \textbf{zero-shot} language for \diffur~models --- no Gujarati paraphrase data was seen during training. We see that while the vanilla \diffur~model performs poorly, the \diffurindic~is competitive with baselines and the \exemp~variant significantly outperforms other systems. We also see that the \exemp~variant lead to better style transfer control (bottom plot, closer to $x=1$ is better), giving style variation with $\lambda$ without loss in semantics (X-axis). }
    \label{fig:gujarati-test-formality-eval-plots}
\end{figure*}

A full breakdown of results by individual metrics, along with plots showing variation with change in $\lambda$, is provided for --- Hindi (\tableref{tab:hindi-test-formality-eval}, \figureref{fig:hindi-test-formality-eval-plots}), Bengali (\tableref{tab:bengali-test-formality-eval}, \figureref{fig:bengali-test-formality-eval-plots}), Kannada (\tableref{tab:kannada-test-formality-eval}, \figureref{fig:kannada-test-formality-eval-plots}), Telugu (\tableref{tab:telugu-test-formality-eval}, \figureref{fig:telugu-test-formality-eval-plots}), Gujarati (\tableref{tab:gujarati-test-formality-eval}, \figureref{fig:gujarati-test-formality-eval-plots}).

\section{More Model Outputs}
\label{appendix:model-outputs}

\begin{figure*}[t!]
    \centering
    \includegraphics[width=0.99\textwidth]{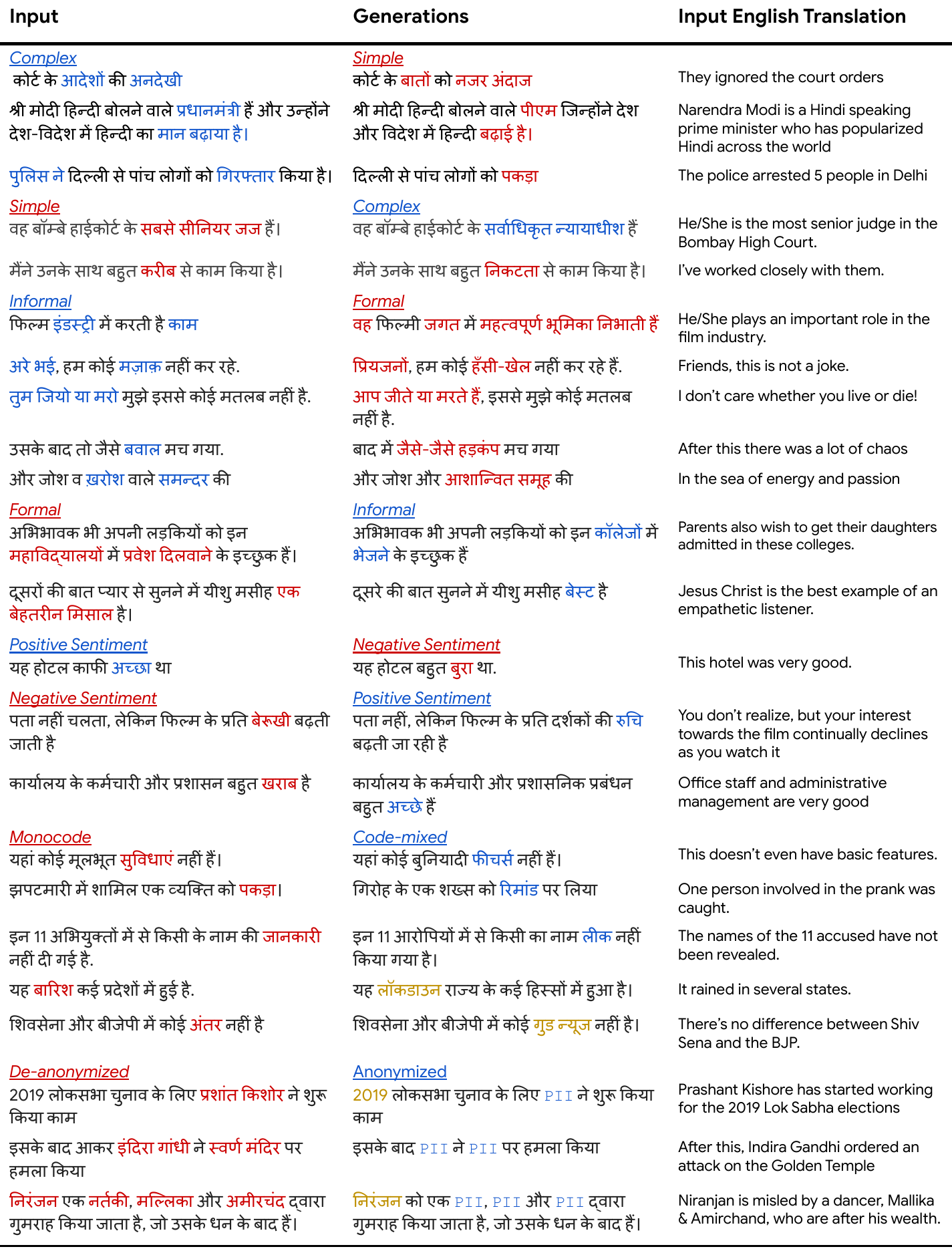}
    \caption{More qualitative examples of generations from our system (see \figureref{fig:qualitative} for main table with qualitative analysis). Red and blue colours indicate attribute-specific features, while golden text represents model errors.}
    \label{fig:qualitative-all}
\end{figure*}

Please refer to \figureref{fig:qualitative-all}. In the main body, \figureref{fig:qualitative} has a few examples as well with detailed analysis.

\section{Paraphrase Diversity}
\label{appendix:paraphrase-diversity}

\begin{figure*}[t!]
    \centering
    \includegraphics[width=0.49\textwidth]{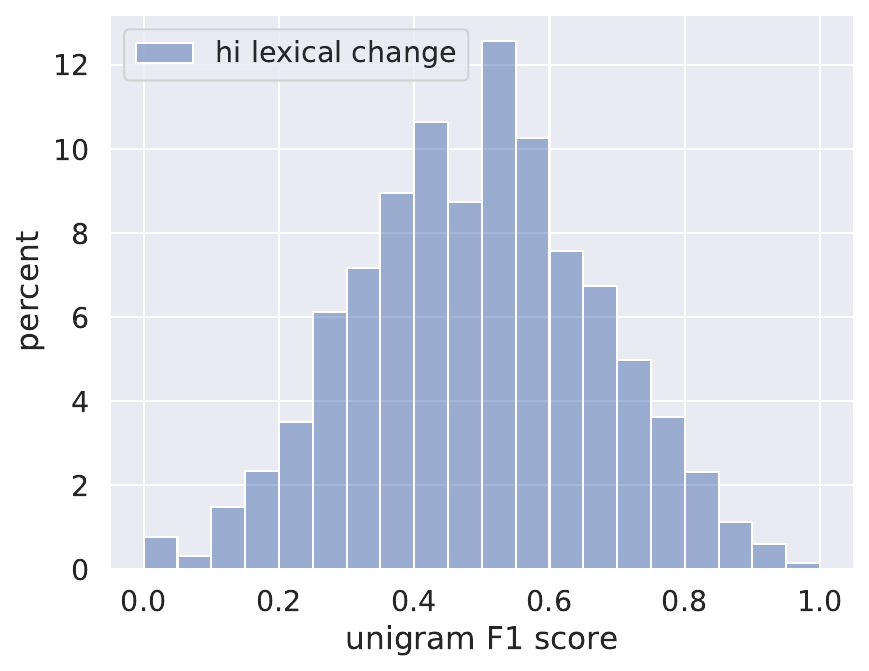}
    \includegraphics[width=0.49\textwidth]{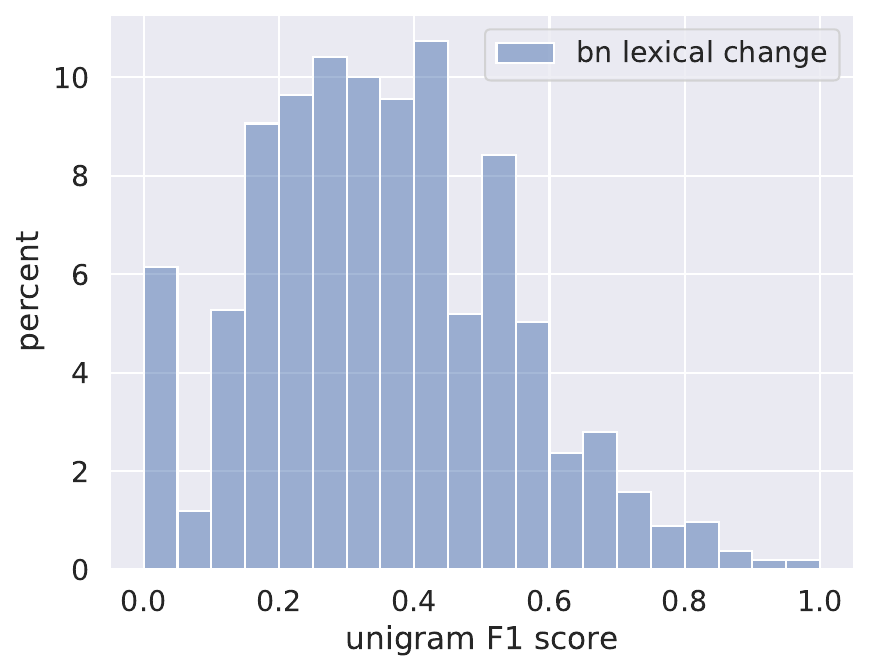}
     \includegraphics[width=0.49\textwidth]{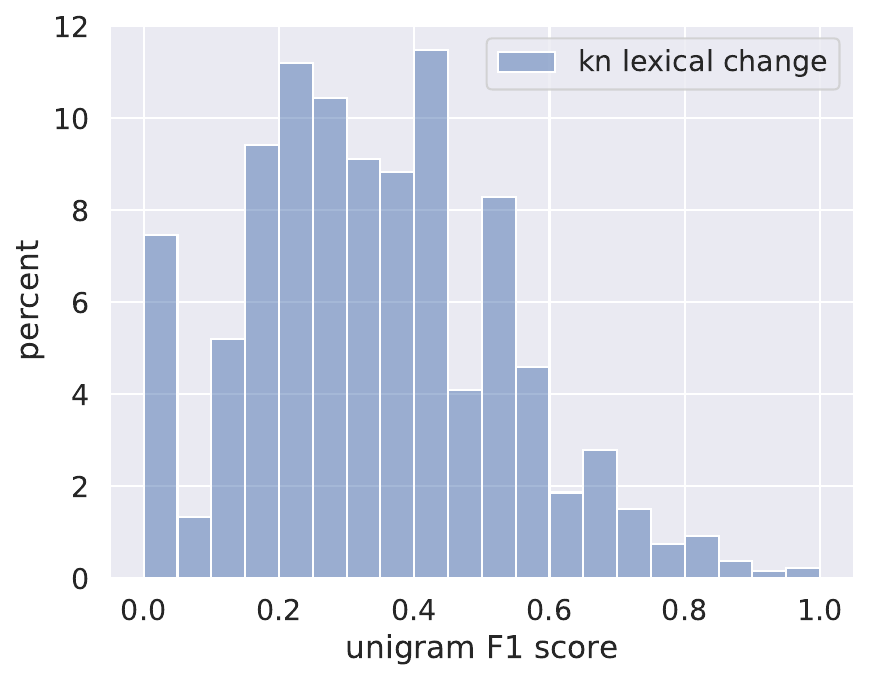}
      \includegraphics[width=0.49\textwidth]{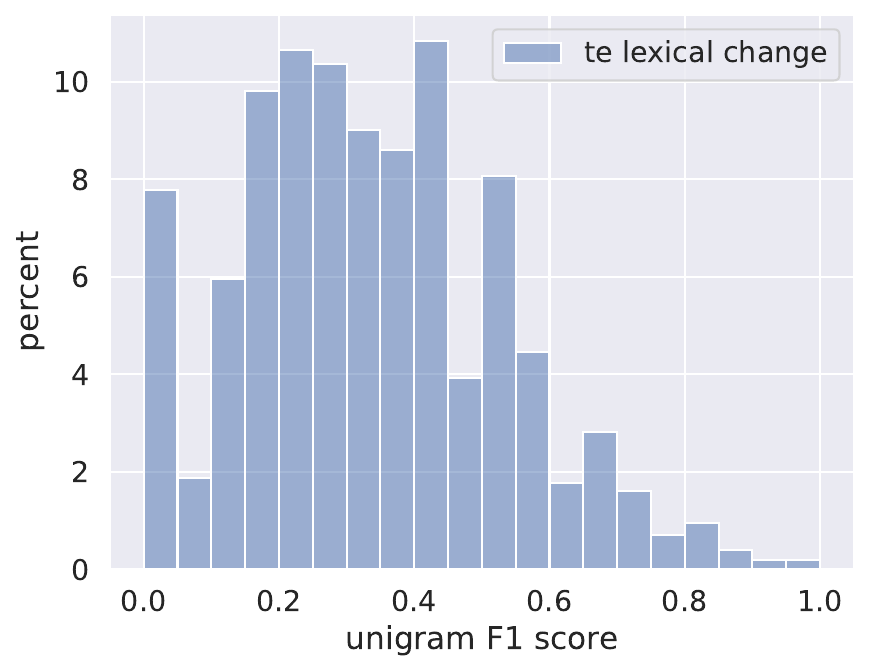}
       \includegraphics[width=0.49\textwidth]{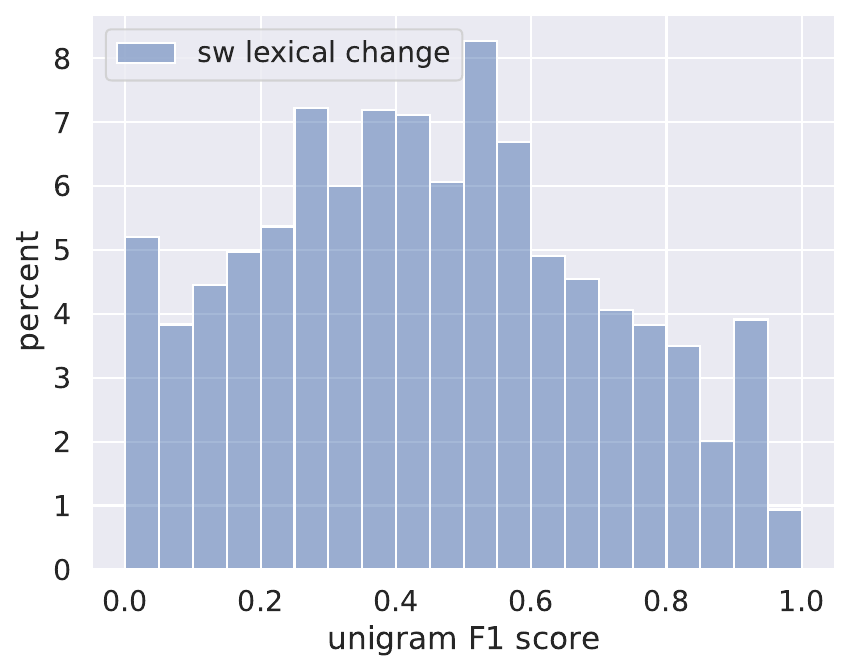}
        \includegraphics[width=0.49\textwidth]{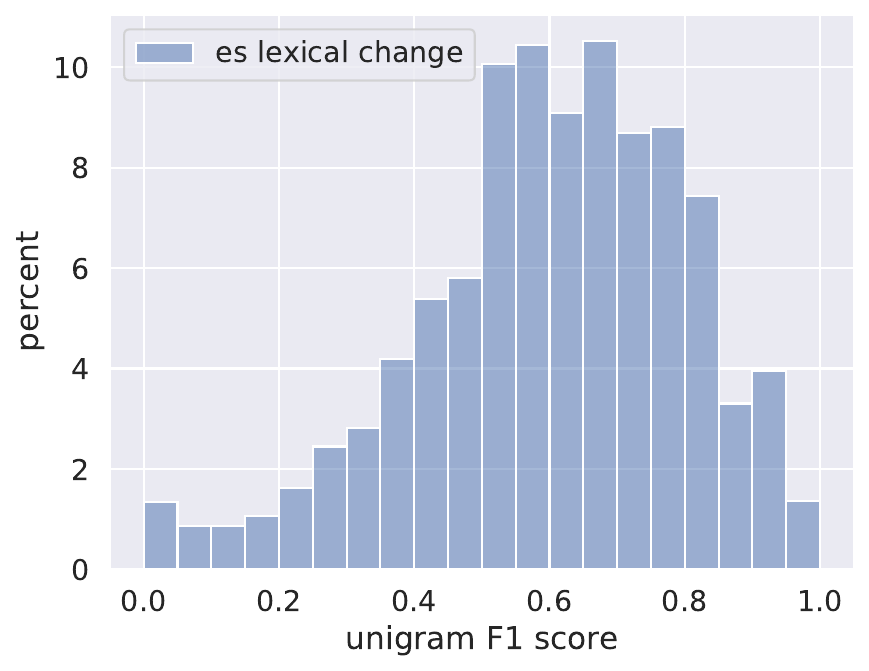}
    \caption{Lexical overlap between paraphrases used in our \diffur~training strategy for six different languages (Hindi, Bengali, Kannada, Telugu, Swahili and Spanish). The wide spread of the histogram and sufficient percentage of low overlap pairs confirm the lexical diversity of the paraphrases used. The lexical overlap is measured using the unigram F1 score, using the implementation from the SQuAD evaluation script~\citep{rajpurkar-etal-2016-squad}.}
    \label{fig:lex-overlap-paraphrases}
\end{figure*}

In \figureref{fig:lex-overlap-paraphrases} we measure the lexical overlap between paraphrases used in our \diffur~training strategy for six different languages (Hindi, Bengali, Kannada, Telugu, Swahili and Spanish). The lexical overlap is measured using the unigram F1 score, using the implementation from the SQuAD evaluation script~\citep{rajpurkar-etal-2016-squad}. The wide spread of the histogram and sufficient percentage of low overlap pairs confirm the lexical diversity of the paraphrases used. As shown in prior work~\citep{krishna-etal-2020-reformulating}, high lexical diversity of paraphrases is helpful for changing the input style.

\begin{table*}[t!]
\small
\begin{center}
\begin{tabular}{ l|rrr|rrr|rrr|rr| } 
 \toprule
 Model & \multicolumn{3}{c|}{$\lambdamax/3$} & \multicolumn{3}{c|}{$2\lambdamax/3$} & \multicolumn{3}{c|}{$\lambdamax$} & \multicolumn{2}{c|}{Overall}\\
  & $\lambda$ & \relagg & \stylechange &  $\lambda$ & \relagg & \stylechange &  $\lambda$ & \relagg & \stylechange & \stylecalib & \stylecalibinp \\
 \midrule
 \ur~\shortcite{garcia2021towards} & 0.5 & 22.1 & 5.2 & 1.0 & 26.9 & 8.9 & 1.5 & 30.4 & 18.7 & 29.2 & 31.6 \\
 \urindic & 0.5 & 53.2 & 13.4 & 1.0 & 58.3 & 18.8 & 1.5 & 54.6 & 26.7 & 60.7 & 65.1 \\
  \ur~+ \textsc{bt} & 0.3 & 53.2 & 21.4 & 0.7 & 53.9 & 23.5 & 1.0 & 49.1 & 26.9 & 43.4 & 58.8  \\
 \urindic~+ \textsc{bt} & 0.3 & 57.3 & 22.9 & 0.7 & 59.4 & 24.6 & 1.0 & 60.0  &  26.7 & 38.7 & 56.0 \\
 \diffur & 0.5 & 65.8 & 16.6 & 1.0 & 71.1 & 26.0 & 1.5 & 67.1 & 21.9 & 64.9 & 72.5 \\
 \diffurindic & 0.8 & 67.2 & 17.9 & 1.7 & 72.6 & 27.3 & 2.5 & 65.0 & \textbf{36.7} & \textbf{69.6} & \textbf{75.5} \\
 \exemp & 0.8 & 56.6 & 11.3 & 1.7 & 72.6 & 18.1 & 2.5 & 78.1 & 29.9 & 69.0 & 71.8 \\
\bottomrule
\end{tabular}
\end{center}
\vspace{-0.1in}
\caption{Evaluation of extent to which the magnitude of hindi formality transfer can be controlled with $\lambda$. We find that \diffurindic, \exemp~are best at calibrating style change to input $\lambda$ (\stylecalib, \stylecalibinp), giving the higher style score increase (\stylechange) at $\lambda=\lambdamax$ (details of evaluation setup and metrics in \sectionref{sec:evaluating-control}, \appendixref{appendix:more-control-evals}).}

\label{tab:hindi-formality-eval-test-control-full}
\end{table*}



\begin{table*}[t!]
\begin{center}
\begin{tabular}{ lrrrrr|rr } 
 \toprule
 Ablation & \copymetric $(\downarrow)$ & \lang & \simmetric & \relacc & \absacc & \relagg & \absagg \\
 \midrule
 \diffurindic~(hindi only) & 2.0 & 97.0 & 78.4 & 89.8 & 39.7 & 67.3 & 24.6 \\
- no paraphrase** & 21.0 & 98.3 & 92.2 & 60.0 & 15.7 & 51.9 & 10.7 \\
- no paraphrase ($p, \lambda =0.6, 3$) & 14.2 & 98.7 & 81.0 & 70.9 & 28.1 & 51.6 & 12.5 \\
- no paraphrase semantic filtering & 2.2 & 97.2 & 72.2 & 89.1 & 38.6 & 60.7 & 19.6 \\
- no vector differences** & 0.0 & 54.3 & 3.2 & 99.0 & 90.0 & 2.4 & 1.0 \\
- no vector differences ($\lambda = 0.5$) & 0.9 & 97.4 & 66.8 & 86.4 & 36.5 & 53.5 & 17.3 \\
- mC4 instead of Samanantar & 1.5 & 91.4 & 82.0 & 89.3 & 39.0 & 67.7 & 24.2 \\
- mC4 + exemplar instead of target & 5.5 & 23.8 & 82.3 & 77.2 & 32.3 & 13.8 & 3.2 \\
\bottomrule
\end{tabular}
\end{center}
\caption{Ablation study on Hindi formality transfer validation set using beam size of 4 and $\lambda = 2.0$ unless the optimal hyperparameters were different (marked by **). As shown by the overall \absagg~scores, removing any component of our design leads to an overall performance drop, sometimes significantly. For a detailed description of analysis and results, see \appendixref{appendix:ablation-explanations}. For detailed metric descriptions, see \sectionref{sec:evaluation}.}
\label{tab:hindi-formality-ablations}
\end{table*}

\begin{table*}[t!]
\begin{center}
\begin{tabular}{ lrrrrrr|rr } 
 \toprule
 Decoding & \copymetric $(\downarrow)$ & \fmetric $(\downarrow)$ & \lang & \simmetric & \relacc & \absacc & \relagg & \absagg \\
 \midrule
 beam 4 & 1.8 & 52.7 & 95.8 & 73.3 & 94.7 & 51.6 & \textbf{66.2} & \textbf{32.3} \\
 beam 1 & 1.2 & 47.4 & 92.3 & 61.7 & 95.7 & 62.5 & 55.8 & 31.4 \\
 \midrule
 top-$p$ 0.6 & 1.0 & 45.3 & 91.5 & 56.6 & 96.2 & 65.9 & 51.3 & 29.9 \\
 top-$p$ 0.75 & 0.9 & 43.1 & 90.3 & 52.4 & 96.3 & 69.0 & 47.3 & 28.2 \\
 top-$p$ 0.9 & 0.7 & 40.4 & 89.4 & 46.8 & 96.6 & 71.7 & 42.4 & 26.5 \\
\bottomrule
\end{tabular}
\end{center}
\caption{Automatic evaluation of different decoding algorithms (top-$p$ sampling and beam search) on the \exemp~model for Hindi formality transfer (validation set) using $\lambda=3.0$. As expected, output diversity (\fmetric, \copymetric) and style accuracy (\relacc, \absacc) improves as we move down the table, but compromise semantic preservation (\simmetric), bringing the overall performance (\relagg, \absagg) down. Also see \figureref{fig:ablation-decoding-graphs} for a comparison across $\lambda$ values, and \sectionref{sec:evaluation} for detailed metric descriptions.}
\label{tab:hindi-formality-eval-decoding-scheme}
\end{table*}

\begin{table*}[t!]
\small
\begin{center}
\begin{tabular}{ lrrrrrrrrrrrr } 
 \toprule
 Model & \multicolumn{2}{c}{Hindi} & \multicolumn{2}{c}{Bengali} & \multicolumn{2}{c}{Kannada} & \multicolumn{2}{c}{Telugu} & \multicolumn{2}{c}{Gujarati} &  \\
 & \relagg & \absagg  & \relagg & \absagg  & \relagg & \absagg  & \relagg & \absagg  & \relagg & \absagg\\
 \midrule
 \ur~\shortcite{garcia2021towards} & 34.5 & 13.4 & 33.8 & 9.0 & 26.8 & 8.8 & 24.3 & 10.7 & 25.6 & 5.9 \\
 \ur~+ \textsc{bt} & 61.6 & 24.2 & 65.6 & 22.8 & 48.8 & 16.0 & 48.7 & 17.6 & 56.3 & 15.1 \\
 \diffur & 79.4 & 30.3 & 81.7 & 36.0 & 79.0 & 43.4 & 79.7 & 38.0 & 0.5 & 0.2 \\
 \urindic & 62.0 & 23.9 & 69.3 & 29.3 & 64.6 & 22.2 & 65.0 & 25.8 & 59.0 & 13.8 \\
 \urindic~+ \textsc{bt} & 68.0 & 28.1 &  73.5 & 33.3 & 72.6 & 29.7 & 71.6 & 31.4 & 68.4 & 21.7 \\
 \diffurindic & 80.0 & 32.4 & 80.0 & 32.3 & 79.9 & 41.4 & 78.8 & 37.0 & 38.9 & 16.2 \\
 \exemp & 85.8 & 45.2 & 86.0 & 48.3 & 86.9 & 54.4 & 86.1 & 51.7 & 78.8 & 41.3 \\
\bottomrule
\end{tabular}
\end{center}
\caption{Test set performance across languages for a \textbf{smaller LaBSE semantic similarity threshold} of 0.65. Due to the more relaxed threshold, absolute numbers compared to \tableref{tab:formality-eval-test-multi} are higher. Trends remain similar, with the \diffur~ and \textsc{indic} variants outperforming other competing methods.}
\label{tab:formality-eval-test-multi-labse-lower}
\end{table*}

\begin{table*}[t!]
\small
\begin{center}
\begin{tabular}{ lrrrrrrrrrrrr } 
 \toprule
 Model & \multicolumn{2}{c}{Hindi} & \multicolumn{2}{c}{Bengali} & \multicolumn{2}{c}{Kannada} & \multicolumn{2}{c}{Telugu} & \multicolumn{2}{c}{Gujarati} &  \\
 & \relagg & \absagg  & \relagg & \absagg  & \relagg & \absagg  & \relagg & \absagg  & \relagg & \absagg\\
 \midrule
 \ur~\shortcite{garcia2021towards} & 24.2 & 6.6 & 24.2 & 4.8 & 21.5 & 6.0 & 19.1 & 5.8 & 19.4 & 3.6 \\
 \ur~+ \textsc{bt} &  40.0 & 10.7 & 31.7 & 8.1 & 21.2 & 5.1 & 19.1 & 4.8 & 26.1 & 4.4 \\
 \diffur & 57.1 & 13.0 & 59.6 & 13.0 & 54.5 & 13.8 & 52.8 & 12.8 & 0.2 & 0.0 \\
 \urindic & 49.6 & 13.1 & 54.6 & 12.7 & 50.0 & 11.4 & 48.1 & 11.2 & 45.9 & 6.8 \\
 \urindic~+ \textsc{bt} & 43.7 & 12.9 & 33.9 & 10.2 & 31.9 & 7.8 & 29.4 & 7.8 & 34.0 & 7.4 \\
 \diffurindic & 59.2 & 14.9 & 63.8 & 15.6 & 58.9 & 16.1 & 55.2 & 14.4 & 31.7 & 8.0 \\
 \exemp & 64.8 & 17.9 & 69.8 & 22.0 & 69.3 & 23.5 & 67.5 & 20.6 & 64.0 & 18.2 \\
\bottomrule
\end{tabular}
\end{center}
\caption{Test set performance across languages for a \textbf{larger LaBSE semantic similarity threshold} of 0.85. Due to the stricter threshold, absolute numbers compared to \tableref{tab:formality-eval-test-multi} are lower, however trends are similar, with the \diffur~ and \textsc{indic} variants outperforming other competing methods.}
\label{tab:formality-eval-test-multi-labse-higher}
\end{table*}


\begin{table*}[t!]
\small
\begin{center}
\begin{tabular}{ lrrrrrrr|rr } 
 \toprule
 Model & $\lambda$ & \copymetric $(\downarrow)$ & \fmetric $(\downarrow)$ & \lang & \simmetric & \relacc & \absacc & \relagg & \absagg \\
 \midrule
 \ur~\citep{garcia2021towards} & 1.5 & 45.4 & 77.5 & 98.0 & 84.8 & 45.8 & 22.9 & 30.4 & 10.4 \\
 \urindic  & 1.0 & 10.4 & 70.7 & 95.0 & 93.8 & 67.2 & 23.3 & 58.3 & 18.6 \\
 \midrule
 \ur~+ \textsc{bt} & 0.5 & 0.8 & 44.2 & 92.9 & 85.2 & 72.3 & 27.8 & 54.2 & 17.8 \\
 \urindic~+ \textsc{bt} & 1.0 & 1.1 & 49.5 & 95.9 & 85.1 & 76.3 & 33.1 & 60.0 & 22.2 \\
 \midrule
 \diffur & 1.0 & 4.7 & 61.6 & 97.7 & 89.7 & 82.4 & 31.0 & 71.1 & 22.9  \\
 \diffurindic &  1.5 & 5.3 & 63.7 & 98.0 & 91.9 & 81.6 & 30.5 & 72.5 & 23.7 \\
 & 2.0 & 3.4 & 57.5 & 98.3 & 84.8 & 86.4 & 36.8 & 70.6 & 24.0 \\
 \exemp & 2.5 & 4.4 & 61.9 & 97.2 & 89.7 & 89.7 & 34.0 & 78.1 & 27.5 \\
 & 3.0 & 2.0 & 52.5 & 95.9 & 72.1 & 94.1 & 51.9 & 64.8 & 32.2 \\
\bottomrule
\end{tabular}
\end{center}
\vspace{-0.05in}
\caption{Performance breakdown of Hindi formality transfer by individual metrics described in \sectionref{sec:evaluation}.}
\vspace{-0.05in}
\label{tab:hindi-test-formality-eval}
\end{table*}

\begin{table*}[t!]
\small
\begin{center}
\begin{tabular}{ lrrrrrrr|rr } 
 \toprule
 Model & $\lambda$ & \copymetric $(\downarrow)$ & \fmetric $(\downarrow)$ &  \lang & \simmetric & \relacc & \absacc & \relagg & \absagg \\
 \midrule
 \textsc{ur}~\citep{garcia2021towards} & 1.5 & 21.5 & 69.1 & 99.9 & 87.3 & 42.4 & 15.6 & 30.4 & 7.2 \\
 \urindic & 1.0 & 4.4 & 58.9 & 99.0 & 95.7 & 69.8 & 19.5 & 65.5 & 17.3 \\
 & 1.5 & 2.4 & 47.5 & 97.6 & 79.8 & 80.0 & 37.4 & 59.6 & 22.3 \\
  \midrule
 \ur~+ \textsc{bt} & 0.5 & 0.2 & 30.4 & 97.8 & 80.6 & 71.8 & 22.3 & 55.6 & 15.0 \\
 & 1.0 & 0.1 & 27.0 & 95.4 & 73.6 & 77.6 & 29.6 & 53.5 & 16.9 \\
 \urindic~+ \textsc{bt} & 1.0 & 0.4 & 34.9 & 99.8 & 80.6 & 78.3 & 31.4 & 61.1 & 22.0\\
  \midrule
  \diffur & 1.0 & 2.1 & 50.6 & 99.9 & 91.6 & 80.8 & 25.2 & 72.7 & 20.9 \\
 & 1.5 & 1.1 & 40.6 & 99.9 & 75.8 & 89.1 & 39.7 & 65.8 & 25.2 \\
 \diffurindic & 1.5 & 2.0 & 53.1 & 99.9 & 94.2 & 80.7 & 24.6 & 75.4 & 21.8 \\
 & 2.5 & 0.9 & 41.4 & 99.9 & 75.6 & 86.1 & 36.9 & 64.6 & 24.3 \\
 \exemp & 2.5 & 1.8 & 49.5 & 99.9 & 91.9 & 87.9 & 39.1 & 80.0 & 33.8 \\
 & 3.0 & 1.0 & 40.0 & 99.1 & 73.0 & 92.1 & 56.5 & 65.3 & 35.0 \\
\bottomrule
\end{tabular}
\end{center}
\caption{Performance breakdown of Bengali formality transfer by individual metrics described in \sectionref{sec:evaluation}.}
\label{tab:bengali-test-formality-eval}
\end{table*}

\begin{table*}[t]
\small
\begin{center}
\begin{tabular}{ lrrrrrrr|rr } 
 \toprule
 Model & $\lambda$ & \copymetric  $(\downarrow)$ & \fmetric $(\downarrow)$ & \lang & \simmetric & \relacc & \absacc & \relagg & \absagg \\
 \midrule
 \ur~\citep{garcia2021towards} & 1.5 & 52.0 & 86.8 & 99.9 & 95.0 & 29.9 & 11.2 & 25.5 & 8.0 \\
  \urindic & 1.0 & 8.6 & 62.9 & 98.3 & 94.5 & 67.0 & 20.8 & 61.3 & 17.8\\
  \midrule
 \ur~+ \textsc{bt} & 0.5 & 0.3 & 26.0 & 77.8 & 75.5 & 67.2 & 23.3 & 39.8 & 11.9  \\
 \urindic~+ \textsc{bt} & 0.5 & 1.6 & 40.6 & 99.9 & 82.3 & 73.9 & 26.8 & 59.2 & 19.1 \\
  & 1.0 & 1.4 & 37.7 & 99.8 & 76.8 & 78.3 & 32.8 & 58.1 & 21.0\\
  \midrule
    \diffur & 1.0 & 3.0 & 47.4 & 99.8 & 87.9 & 80.3 & 30.5 & 69.2 & 23.6 \\
   & 2.0 & 2.2 & 39.6 & 99.9 & 73.0 & 87.8 & 48.3 & 62.1 & 29.1 \\
 \diffurindic & 1.5 & 2.9 & 50.3 & 99.9 & 91.5 & 81.2 & 32.2 & 73.1 & 26.4 \\
 & 2.0 & 2.3 & 45.2 & 99.9 & 82.7 & 85.1 & 42.3 & 68.5 & 29.3 \\
 \exemp & 2.0 & 5.4 & 59.6 & 100 & 97.5 & 82.9 & 28.9 & 80.4 & 27.5\\
 & 3.0 & 2.1 & 42.7 & 99.1 & 71.7 & 92.6 & 63.4 & 64.5 & 39.4 \\
\bottomrule
\end{tabular}
\end{center}
\caption{Performance breakdown of Kannada formality transfer by individual metrics described in \sectionref{sec:evaluation}.}
\label{tab:kannada-test-formality-eval}
\end{table*}

\begin{table*}[t]
\small
\begin{center}
\begin{tabular}{ lrrrrrrr|rr } 
 \toprule
 Model & $\lambda$ & \copymetric $(\downarrow)$ & \fmetric $(\downarrow)$ & \lang & \simmetric & \relacc & \absacc & \relagg & \absagg \\
 \midrule
 \ur~\shortcite{garcia2021towards} & 1.5 & 51.3 & 87.0 & 100 & 96.3 & 26.3 & 10.1 & 22.8 & 7.5  \\
 & 2.0 & 35.0 & 68.2 & 99.9 & 73.0 & 45.4 & 28.6 & 20.7 & 8.4 \\
 \urindic & 1.0 & 10.4 & 64.5 & 98.8 & 94.3 & 65.6 & 20.2 & 59.8 & 16.7 \\
 & 1.5 & 5.9 & 53.5 & 97.3 & 80.0 & 74.9 & 33.1 & 55.9 & 19.9 \\
 \midrule
  \ur~+ \textsc{bt} & 0.5 & 0.2 & 26.3 & 82.4 & 73.4 & 65.6 & 23.4 & 38.4 & 11.3 \\
  & 1.0 & 0.1 & 19.8 & 74.9 & 64.7 & 71.2 & 31.6 & 33.1 & 11.6 \\
  \urindic~+ \textsc{bt} & 0.5 & 0.6 & 39.2 & 99.9 & 79.6 & 73.5 & 26.2 & 56.8 & 17.9  \\
 & 1.0 & 0.5 & 36.1 & 99.7 & 74.0 & 78.5 & 35.9 & 56.0 & 22.2 \\
 \midrule
  \diffur & 1.0 & 1.7 & 46.0 & 99.9 & 87.9 & 80.5 & 27.6 & 69.4 & 21.5 \\
  & 2.5 & 0.9 & 36.0 & 99.8 & 68.4 & 90.2 & 47.2 & 59.9 & 27.1 \\
 \diffurindic & 1.0 & 2.4 & 50.1 & 99.9 & 91.7 & 78.7 & 28.7 & 71.0 & 23.7 \\
 & 1.5 & 1.4 & 44.6 & 99.9 & 83.6 & 83.6 & 38.4 & 68.2 & 27.1 \\
 \exemp & 2.0 & 3.8 & 55.8 & 99.9 & 95.7 & 84.0 & 31.2 & 79.8 & 28.6 \\ 
 & 2.5 & 1.8 & 47.0 & 99.5 & 85.8 & 90.1 & 48.4 & 76.0 & 37.9 \\
\bottomrule
\end{tabular}
\end{center}
\caption{Performance breakdown of Telugu formality transfer by individual metrics described in \sectionref{sec:evaluation}.}
\label{tab:telugu-test-formality-eval}
\end{table*}

\begin{table*}[t]
\small
\begin{center}
\begin{tabular}{ lrrrrrrr|rr } 
 \toprule
 Model & $\lambda$ & \copymetric $(\downarrow)$ & \fmetric $(\downarrow)$ & \lang & \simmetric & \relacc & \absacc & \relagg & \absagg \\
 \midrule
 \ur~\shortcite{garcia2021towards} & 1.5 & 62.6 & 89.1 & 99.9 & 93.1 & 30.2 & 9.3 & 23.7 & 5.0 \\
  \urindic & 1.0 & 17.5 & 73.6 & 98.4 & 96.8 & 57.6 & 11.7 & 54.0 & 9.9 \\
 & 1.5 & 10.9 & 62.7 & 96.9 & 85.4 & 67.0 & 19.2 & 53.0 & 10.7 \\
 \midrule
  \ur~+ \textsc{bt} & 0.5 & 0.5 & 34.3 & 87.3 & 77.6 & 69.1 & 17.8 & 46.3 & 9.8 \\
  & 1.0 & 0.3 & 26.5 & 78.8 & 67.6 & 74.8 & 27.2 & 39.1 & 10.4\\
   \urindic~+ \textsc{bt} & 0.5 & 1.9 & 47.4 & 99.9 & 87.1 & 68.1 & 22.0 & 57.7 & 16.8 \\
  \midrule
  \diffur & 0.5 & 0.0 & 5.7 & 1.2 & 81.3 & 73.2 & 25.7 & 0.4 & 0.2 \\
 \diffurindic & 0.5 & 1.1 & 34.7 & 54.9 & 95.6 & 68.6 & 18.6 & 37.4 & 9.0  \\
 & 1.5 & 0.4 & 24.2 & 46.0 & 74.7 & 78.5 & 40.0 & 29.2 & 13.0 \\
 \exemp & 2.0 & 7.7 & 65.4 & 98.6 & 96.2 & 79.3 & 25.0 & 75.0 & 22.3 \\
 & 2.5 & 4.5 & 54.6 & 95.1 & 85.5 & 86.0 & 45.8 & 69.8 & 33.1 \\
\bottomrule
\end{tabular}
\end{center}
\caption{Performance breakdown of Gujarati formality transfer by individual metrics described in \sectionref{sec:evaluation}.}
\label{tab:gujarati-test-formality-eval}
\end{table*}

\end{document}